\documentclass[runningheads]{llncs}

\usepackage[T1]{fontenc}
\usepackage{graphicx}
\usepackage{xcolor}
\usepackage{mathtools}
\usepackage{comment}
\usepackage{cancel} 
\usepackage{amssymb}
\usepackage{mathtools}
\usepackage{amsmath,bm}

\newcommand{\ms}[1]{[\![#1]\!]}

\begin{document}
\title{A Data-Transparent Probabilistic Model of Temporal Propositional Abstraction}
\titlerunning{A Data-Transparent Probabilistic Model}
\author{Hiroyuki Kido\orcidID{00000-0002-7622-4428}}
\authorrunning{H. Kido}
\institute{Cardiff University, Park Place, CF10 3AT, Cardiff, UK}
\maketitle              % typeset the header of the contribution
\begin{abstract}
Standard probabilistic models face fundamental challenges such as data scarcity, a large hypothesis space, and poor data transparency. To address these challenges, we propose a novel probabilistic model of data-driven temporal propositional reasoning. Unlike conventional probabilistic models where data is a product of domain knowledge encoded in the probabilistic model, we explore the reverse direction where domain knowledge is a product of data encoded in the probabilistic model. This more data-driven perspective suggests no distinction between maximum likelihood parameter learning and temporal propositional reasoning. We show that our probabilistic model is equivalent to a highest-order, i.e., full-memory, Markov chain, and our model requires no distinction between hidden and observable variables. We discuss that limits provide a natural and mathematically rigorous way to handle data scarcity, including the zero-frequency problem. We also discuss that a probability distribution over data generated by our probabilistic model helps data transparency by revealing influential data used in predictions. The reproducibility of this theoretical work is fully demonstrated by the included proofs.
\end{abstract}
%
%%%%%%%%%%%%%%%%%%%%%%%%%%%%%%%%%%%%%%%%%%%%%%%%%%%%%%%%%%%%%%%%%%%%%%%%%%%%%%%%%%%%%%%%%%%%%%%%%%%%%%%%%%%%%%%%%%%%%%%%%%%%%%%%%%%%%%%%%%%%%%%%%%%%%%%%%%%%%%%%%%%%%%%%%%%%%%%%%%%%%%%%%%%%%%%%%%%%%%%%%%%%%%%%%%%%%%%%%%%%%%%%%%%%%%%%%%%%%%%%%%%%%%%%%%%%%%%%%%%%%%%%%%%%%%%%%%%%%%%%%%%%%%%%%%%%%%%%%%%%%%%%%%%%%%%%%%%%%%%%%%%%%%%%%%%%%%%%%%%%%%%%%%%%%%%%%%%%%%%%%%%%%%%%%%%%%%%%%%%%%%%%%%%%%%%%%%%%%%%%%%%%%%%%%%%%%%%%%%%%%%%%%%%%
\section{Introduction}
Probability theory underlies modern AI \cite{Russell:20}. Probabilistic modelling has led to various successful AI applications, such as computer vision, speech recognition, and natural language processing \cite{Pearl:88,Bishop:06}. However, it inherently involves fundamental challenges such as data scarcity, an exponentially growing hypothesis space, and poor data transparency. To illustrate these challenges, let us consider the following simple, discrete-time, discrete-state localisation problem.
%%%
\begin{figure}[t]
\begin{center}
\includegraphics[scale=0.25]{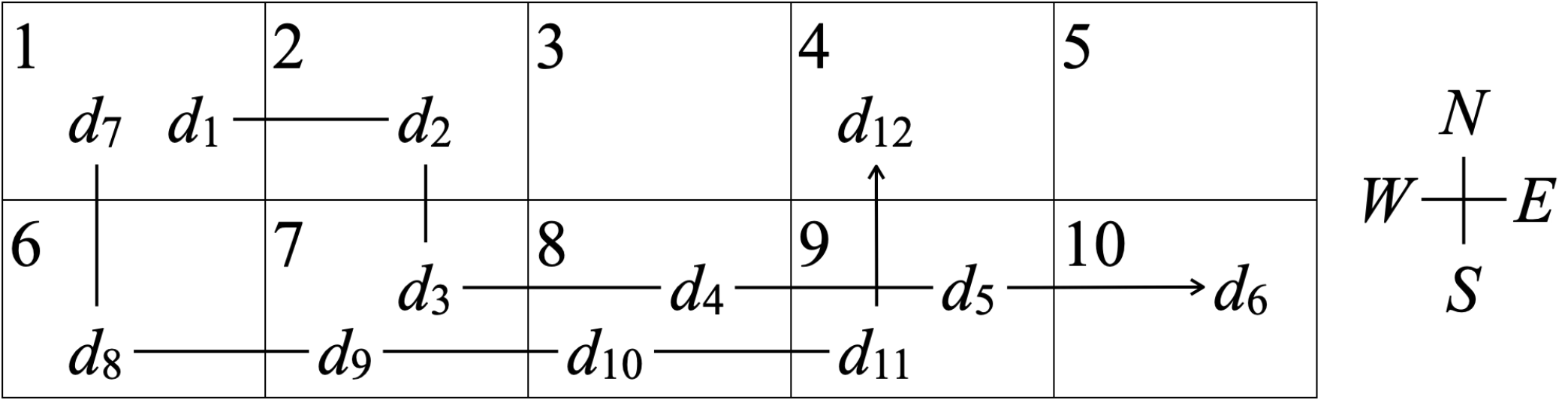}
\caption{Twelve data collected by a robot exploring a building with ten rooms.}
\label{fig:maze}
\end{center}
\end{figure}
%%%
\begin{example}\label{ex:intro}
Figure \ref{fig:maze} shows a building with ten rooms. The room number is shown in the northwest corner of each room. The two arrows indicate the tracks of a robot, and $d_{k}$ denotes the data collected by the robot in the room, for all $k\in\{1,2,...,12\}$. Using the twelve data, we want to find the location of the robot exploring the building. Suppose that the robot moved through Rooms 2, 3, and 8. Where is the robot likely to be two time steps after Room 8?
\qed
\end{example}
%%%
The standard approaches to this problem are probabilistic modelling \cite{Bishop:06,Russell:20} such as Markov chains and hidden Markov models \cite{Rabiner:89,Mor:21}. However, they are not fundamentally free from the following issues.
\par
The first issue is data scarcity. Since the robot does not experience Room 3, a probabilistic model naively trained with the twelve data cannot predict the robot location due to zero frequency. While data smoothing \cite{Murphy:12,Russell:20} mitigates the problem, it is effective only when the number of parameters in the probabilistic model is sufficiently small.
\par
The second issue is an exponentially growing hypothesis space. The three data beginning with $d_{2}$ best match the known past robot locations. Indeed, assuming $d_{2}$ corresponds to time step 1, denoted by Time 1, the series of data correctly explains the robot locations at Times 1 and 3. The robot is then predicted to be in Room 10 at Time 5. While this idea seems promising, it cannot be easily generalised using probabilistic models. The number of parameters in an $n$th-order Markov chain with $r$ states is $(r-1)\times r^{n}$. Thus, even this simple problem requires $90,000$ (i.e., $9\times 10^{4}$) parameters in a 4th-order Markov chain, which takes into account all the past four time steps to predict the next one.
\par
The third issue is data transparency. The three data beginning with $d_{8}$ next best match the known past robot locations. Indeed, assuming $d_{8}$ corresponds to Time 1, the series of data correctly explains the robot location at Time 3. The robot is then predicted to be in Room 4, rather than Room 10, at Time 5. Now, the probability of Room 4 should be lower than that of Room 10, considering the consistency with the known robot locations. However, what if the series of data beginning with $d_{8}$ occurs repeatedly? At some point, consistency in quantity may surpass consistency in quality. To the best of our knowledge, however, standard probabilistic models cannot justify this result with reference to actual data such as $d_{2}$ and $d_{8}$. This is because learning is typically the process of exploiting data to adjust the parameters of probabilistic models, whereas reasoning is the process of using the parameters, not the data itself, to make predictions.
\par
In this paper, we propose a novel data-transparent probabilistic model as a simple yet unconventional approach to addressing the aforementioned issues. The key components of the probabilistic model are data, models (i.e., valuations) in propositional logic and propositional formulas $X$ being true, for each time step $t$, denoted by $d^{t}$, $m^{t}$ and $x^{t}$, respectively. We will argue that the probability of $x^{t}$, denoted by $p(x^{t})$, should be given as follows.
\begin{eqnarray}\label{eq:intro}
&&p(x^{t})=\sum_{m^{t}}\sum_{d^{1}}\sum_{d^{2}}\cdots\sum_{d^{t}}p(x^{t}, m^{t}, d^{1}, d^{2}, ..., d^{t})\text{ where}\nonumber\\
&&p(x^{t}, m^{t}, d^{1}, d^{2},..., d^{t})=p(x^{t}|m^{t})p(m^{t}|d^{t})p(d^{t}|d^{t-1})p(d^{t-1}|d^{t-2})\cdots p(d^{1})
\end{eqnarray}
Here, the first line is an application of a valid rule of probability theory. The second line is an application of the probabilistic model we formulate in this paper. We will define $p(x^{t}|m^{t})$ based on whether the propositional formula $X$ is true in the model $m^{t}$ at Time $t$, $p(m^{t}|d^{t})$ based on whether the data $d^{t}$ supports the model $m^{t}$ at Time $t$, and $p(d^{t}|d^{t-1})$ based on whether the data $d^{t-1}$ changes to $d^{t}$ at the next time step. In a nutshell, Equation (\ref{eq:intro}) states that the probability of a formula being true depends on whether time-dependent data support a model in which the formula is true. We significantly simplify Equation (\ref{eq:intro}) under the natural assumption that both the data trajectory and the support relation from data to models are deterministic (see Figure \ref{fig:GMs} for an intuitive understanding).
\par
The contributions of this paper are summarised as follows. First, this study is inspired by the inference of abstraction \cite{kido:24-1,kido:24-2}, which suggests logical, statistical, and probabilistic justifications for symbolic reasoning grounded in data. Our probabilistic model additionally incorporates a transition relation between data while maintaining the theoretical justifications and linear computational complexity (see Section \ref{sec:formalism}).
\par
Second, we show that our probabilistic model can be viewed as a highest-order, i.e., most expressive, Markov chain, in which all the past states are used to predict the current state (see Sections \ref{sec:MCs} and \ref{sec:equivalence}). One advantage of our model over Markov chains is data transparency. In our  model, propositional reasoning is fully grounded in data as it always occurs between data and formulas, not between formulas and other formulas (see Section \ref{sec:transparency}). Another advantage is that our model does not require a distinction between observable states and hidden ones (see Section \ref{sec:hidden_states}).
\par
Third, we challenge the conventional view prevailing across AI, cognitive science, and neuroscience that data are assumed to be generated from domain knowledge encoded in probabilistic models, e.g., \cite{lee:03,Itti:09,Hohwy:08,Smith:22,Tenenbaum:06,Lake:15,Lake:17,Dasgupta:20}. Instead, we explore the reverse direction and investigate how domain knowledge can be generated from data, moving toward fully data-driven temporal probabilistic reasoning (see Sections \ref{sec:examples} and \ref{sec:transparency}).
%%%%%%%%%%%%%%%%%%%%%%%%%%%%%%%%%%%%%%%%%%%%%%%%%%%%%%%%%%%%%%%%%%%%%%%%%%%%%%%%%%%%%%%%%%%%%%%%%%%%%%%%%%%%%%%%%%%%%%%%%%%%%%%%%%%%%%%%%%%%%%%%%%%%%%%%%%%%%%%%%%%%%%%%%%%%%%%%%%%%%%%%%%%%%%%%%%%%%%%%%%%%%%%%%%%%%%%%%%%%%%%%%%%%%%%%%%%%%%%%%%%%%%%%%%%%%%%%%%%%%%%%%%%%%%%%%%%%%%%%%%%%%%%%%%%%%%%%%%%%%%%%%%%%%%%%%%%%%%%%
\section{Temporal propositional reasoning from data}\label{sec:formalism}
\subsection{Random variables}
Let $Data=\{d_{1},d_{2},...,d_{K}\}$ be a non-empty set of $K$ data. This set is a multiset, where elements may occur multiple times. For any discrete time $t\in\{1,2,...,T\}$, we assume that $D^{t}$ is a random variable taking values in $Data$. This allows us to handle data that changes over time.
\par
Let $Variables$ be the set of propositional variables, $Values=\{1,0\}$ be the set of truth values meaning true and false, respectively, and $Models=\{m_{1},m_{2},...,m_{L}\}$ be the set of $L$ models, i.e., valuations, in propositional logic. As usual, each model is a function, $Variables\to Values$, that maps each propositional variable to a truth value. For any discrete time $t$, we assume that $M^{t}$ is a random variable taking values in $Models$.
\par
Let ${\cal L}$ be a propositional language. As usual, formulas are constructed from propositional variables using the usual logical connectives such as $\lnot$, $\land$, $\lor$, $\to$, $\leftarrow$, and $\leftrightarrow$. For any discrete time $t$ and propositional formula $X_{i}\in{\cal L}$, $X_{i}^{t}$ is a random variable taking values in $Values$. This allows us to handle the truth values of formulas that vary over time.
\par
In the following sections, we will define the probability distributions over $D^{t}$, $M^{t}$ and $X_{i}^{t}$, denoted by $p(D^{t})$, $p(M^{t})$ and $p(X_{i}^{t})$.
\begin{example}[Continued from Example \ref{eq:intro}]\label{ex:maze}
The problem illustrated in Figure \ref{fig:maze} results in $Data=\{d_{1}, d_{2},...,d_{12}\}$. Let $X_{i}$ be a propositional variable representing that the robot is in Room $i$, for all $i\in\{1,2,...,10\}$. $Models$ then has $2^{10}$ elements, and each model assigns truth values to the ten propositional variables differently. ${X_{1}\to \lnot X_{2}}^{3}$ is a formula representing that `at Time 3, if the robot is in Room 1 then it is not in Room 2.' $X_{1}^{3}\to \lnot X_{2}$ is not a formula as logical connectives can only connect formulas, not time-indexed random variables.
\end{example}
\par
We introduce some abbreviations for readability. $D^{t_{1}:t_{2}}$ denotes the sequence $(D^{t_{1}},D^{t_{1}+1},...,D^{t_{2}})$. The lowercase letter $d^{t}$ denotes a realisation of the random variable $D^{t}$. We often write $D^{t}=d^{t}$ as $d^{t}$ when it is clear from the context. $d^{t_{1}:t_{2}}$ denotes the realisation sequence $(d^{t_{1}},d^{t_{1}+1},...,d^{t_{2}})$. The same argument is applied to the other random variables $M^{t}$ and $X_{i}^{t}$ and their realisations $m^{t}$ and $x_{i}^{t}$. In addition, $X_{i_{1}:i_{2}}^{t_{1}:t_{2}}$ denotes the sequence $(X_{i_{1}}^{t_{1}:t_{2}}$, $X_{i_{1}+1}^{t_{1}:t_{2}}$, ..., $X_{i_{2}}^{t_{1}:t_{2}})$, and $x_{i_{1}:i_{2}}^{t_{1}:t_{2}}$ is the sequence of their realisations. If $t_{1}>t_{2}$ or $i_{1}>i_{2}$ then the sequence is regarded as being empty, and omitted. For example, $p(X_{1}^{1}|D^{1:1},M^{1:1},X_{1:I}^{1:0},X_{1:0}^{1})=p(X_{1}^{1}|D^{1},M^{1})$.
\par
\begin{figure}[t]
\centering
\includegraphics[scale=0.159]{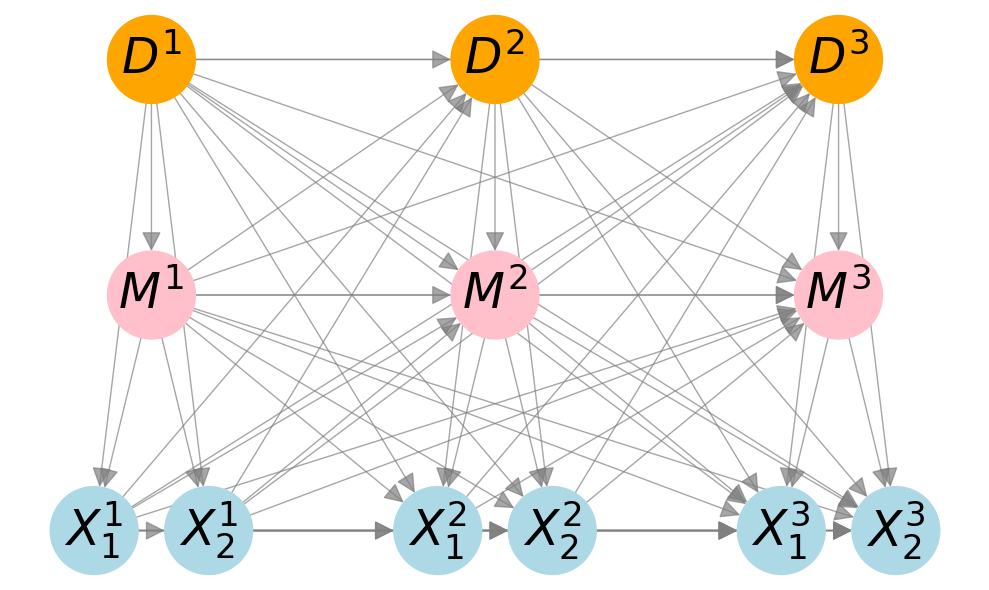}
\includegraphics[scale=0.159]{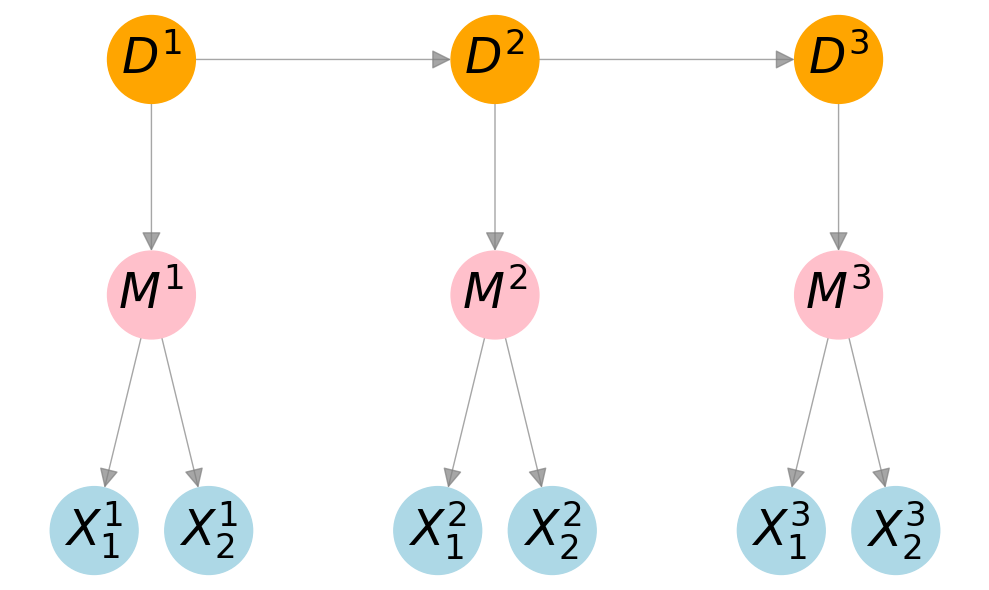}
\includegraphics[scale=0.159]{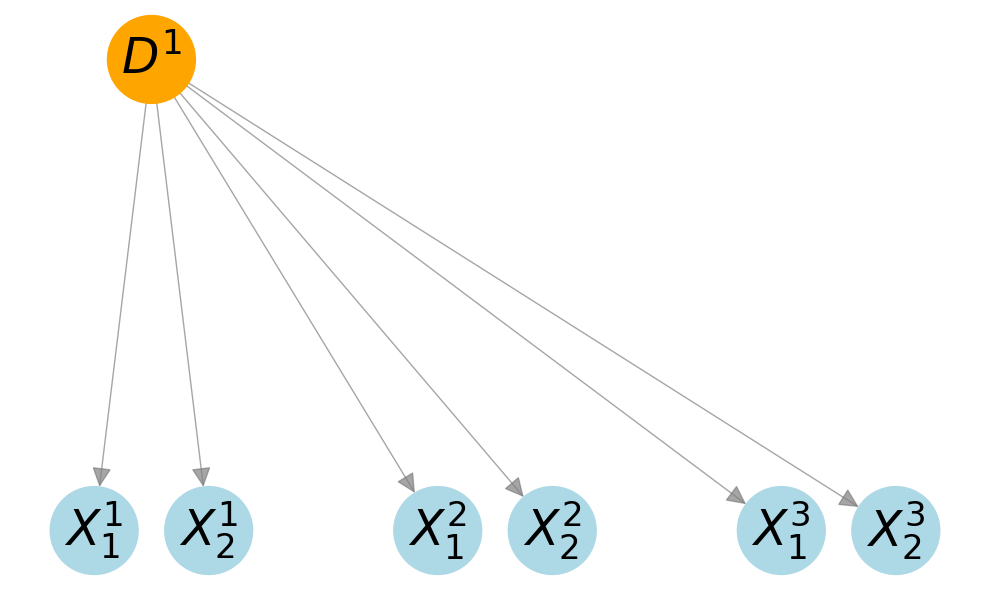}
\caption{This paper shows the equivalence of the three graphical models.}
\label{fig:GMs}
\end{figure}

Now, the full joint distribution over all the introduced random variables can be written as follows using the product rule \cite{Bishop:06} of probability theory.
\begin{eqnarray}
&&p(D^{1:T},M^{1:T},X_{1:I}^{1:T})=\prod_{t=1}^{T}\biggr[p(D^{t}|D^{1:t-1},M^{1:t-1},X_{1:I}^{1:t-1})\biggr.\nonumber\\
&&\biggr.p(M^{t}|D^{1:t},M^{1:t-1},X_{1:I}^{1:t-1})\prod_{i=1}^{I}p(X_{i}^{t}|D^{1:t},M^{1:t},X_{1:I}^{1:t-1},X_{1:i-1}^{t})\biggr]\label{eq:FJD}
\end{eqnarray}
In many cases, we are interested in the marginal distribution over formulas. It can be derived from the full joint distribution using the sum rule \cite{Bishop:06} of probability theory.
\begin{eqnarray}
p(X_{1:I}^{1:T})&=&\sum_{d^{1:T}\in Data^{T}}\sum_{m^{1:T}\in Models^{T}}p(d^{1:T},m^{1:T},X_{1:I}^{1:T})\nonumber\\
&=&\sum_{d^{1:T}\in Data^{T}}\sum_{m^{1:T}\in Models^{T}}\prod_{t=1}^{T}\biggr[p(d^{t}|d^{1:t-1},m^{1:t-1},X_{1:I}^{1:t-1})\biggr.\nonumber\\
&&\biggr.p(m^{t}|d^{1:t},m^{1:t-1},X_{1:I}^{1:t-1})\prod_{i=1}^{I}p(X_{i}^{t}|d^{1:t},m^{1:t},X_{1:I}^{1:t-1},X_{1:i-1}^{t})\biggr]\label{eq:MJD}
\end{eqnarray}
In Figure \ref{fig:GMs}, the leftmost graph represents Equation (\ref{eq:MJD}) with $T=3$ and $I=2$. There is an arrow from each element of the condition to the outcome, for each conditional probability appearing in Equation (\ref{eq:MJD}). Since the graph is a complete directed graph, Equation (\ref{eq:MJD}) states that each random variable can influence each other.
%%%
\begin{example}[Continued from Example \ref{ex:maze}]\label{ex:robot}
What is the probability that the robot is in Room 10 at Time 5 given that it is in Room 2 at Time 1, i.e., $p(X_{10}^{5}=1|X_{2}^{1}=1)$? We simply write it as $p(x_{10}^{5}|x_{2}^{1})$. Using Equation (\ref{eq:MJD}), we have
\begin{eqnarray*}
&&p(x_{10}^{5}|x_{2}^{1})=\frac{p(x_{2}^{1},x_{10}^{5})}{p(x_{2}^{1})}=\frac{\sum_{d^{1:5}\in Data^{5}}\sum_{m^{1:5}\in Models^{5}}\sum_{x_{1:10}^{1:5}\setminus \{x_{2}^{1},x_{10}^{5}\}\in Values^{48}}Z}{\sum_{d^{1:5}\in Data^{5}}\sum_{m^{1:5}\in Models^{5}}\sum_{x_{1:10}^{1:5}\setminus \{x_{2}^{1}\}\in Values^{49}}Z}
\end{eqnarray*}
where $Z$ is given as follows.
\begin{eqnarray*}
Z&=&p(d^{1:5},m^{1:5},x_{1:10}^{1:5})=\prod_{t=1}^{5}\biggr[p(d^{t}|d^{1:t-1},m^{1:t-1},x_{1:10}^{1:t-1})\biggr.\\
&&\biggr.p(m^{t}|d^{1:t},m^{1:t-1},x_{1:10}^{1:t-1})\prod_{i=1}^{10}p(x_{i}^{t}|d^{1:t},m^{1:t},x_{1:10}^{1:t-1},x_{1:i-1}^{t})\biggr]
\end{eqnarray*}
In the next section, we discuss how to simplify the result (see Figure \ref{fig:GMs}).
\qed
\end{example}
%%%%%%%%%%%%%%%%%%%%%%%%%%%%%%%%%%%%%%%%%%%%%%%%%%%%%%%%%%%%%%%%%%%%%%%%%%%%%%%%%%%%%%%%%%%%%%%%%%%%%%%%%%%%%%%%%%%%%%%%%%%%%%%%%%%%%%%%%%%%%%%%%%
\subsection{Data distributions}\label{sec:data_distribution}
We have not yet defined any conditional probabilities appearing in Equation (\ref{eq:FJD}) or (\ref{eq:MJD}). In this section, we define and use the conditional probability of data to simply those equations. To express how data changes over time, we assume a function, $n:Data\to Data$, that maps data at the current time point to data at the next. $|Data|$ denotes the cardinality of $Data$.
%%%
\begin{definition}\label{def:data}
Let $t\in\{1,2,...,T\}$. The conditional probability of $d^{t}$ given $d^{1:t-1}$, $m^{1:t-1}$ and $x_{1:I}^{1:t-1}$ is defined as follows.
\begin{eqnarray*}
p(d^{t}|d^{1:t-1},m^{1:t-1},x_{1:I}^{1:t-1})=
\begin{cases}
\frac{1}{|Data|} & \text{if } t=1\\
1 & \text{if } t\neq 1 \text{ and } d^{t}=n(d^{t-1})\\
0 & \text{otherwise}
\end{cases}
\end{eqnarray*}
\end{definition}
%%%
We derive the following property from Definition \ref{def:data}.
\begin{proposition}\label{prop:data}
Let $t\in\{1,2,...,T\}$. $D^{t}$ is conditionally independent of $D^{1:t-2}$, $M^{1:t-1}$ and $X_{1:I}^{1:t-1}$ given $D^{t-1}$, i.e.,
\begin{eqnarray*}
p(D^{t}|D^{1:t-1},M^{1:t-1},X_{1:I}^{1:t-1})=p(D^{t}|D^{t-1}).
\end{eqnarray*}
\end{proposition}
%%%
\begin{proof}
When $t=1$, both the left- and right-hand sides are $p(D^{1})$. When $t\neq 1$, the right hand side can be expanded as follows for all its realisations.
\begin{eqnarray*}
&&p(d^{t}|d^{t-1})=\frac{p(d^{t-1},d^{t})}{p(d^{t-1})}=\frac{\sum_{d^{1:t-2}}\sum_{m^{1:t-1}}\sum_{x_{1:I}^{1:t-1}}p(d^{1:t},m^{1:t-1},x_{1:I}^{1:t-1})}{p(d^{t-1})}\\
&&=\frac{\sum_{d^{1:t-2}}\sum_{m^{1:t-1}}\sum_{x_{1:I}^{1:t-1}}p(d^{t}|d^{1:t-1},m^{1:t-1},x_{1:I}^{1:t-1})p(d^{1:t-1},m^{1:t-1},x_{1:I}^{1:t-1})}{p(d^{t-1})}
\end{eqnarray*}
Here, the first line is an application of the sum rule, and the second line is an application of the product rule. By definition, the value of $p(d^{t}|d^{1:t-1},m^{1:t-1},x_{1:I}^{1:t-1})$ does not depend on $d^{1:t-2}$, $m^{1:t-1}$ or $x_{1:I}^{1:t-1}$. Therefore, the conditional probability can be moved to the outside of the summations.
\begin{eqnarray*}
&&=\frac{p(d^{t}|d^{1:t-1},m^{1:t-1},x_{1:I}^{1:t-1})\sum_{d^{1:t-2}}\sum_{m^{1:t-1}}\sum_{x_{1:I}^{1:t-1}}p(d^{1:t-1},m^{1:t-1},x_{1:I}^{1:t-1})}{p(d^{t-1})}\\
&&=\frac{p(d^{t}|d^{1:t-1},m^{1:t-1},x_{1:I}^{1:t-1})p(d^{t-1})}{p(d^{t-1})}=p(d^{t}|d^{1:t-1},m^{1:t-1},x_{1:I}^{1:t-1})
\end{eqnarray*}
The second line is an application of the sum rule for marginalisation.
\qed
\end{proof}
%%%
\begin{example}[Continued from Example \ref{ex:robot}]\label{ex:data}
It is clear from Figure \ref{fig:maze} that the probability distribution over data can be directly observed from the environment. For $k\in\{1,2,...,12\}$ and $t\in\{2,3,...,12\}$, 
\begin{eqnarray*}
&&p(D^{1}=d_{k})=\frac{1}{12}\\
&&p(D^{t}=d_{j}|D^{t-1}=d_{i})=
\begin{cases}
1&\text{if }(i,j)\in\{(1,2),(2,3),(3,4),(4,5),(5,6),(6,6),\\
&\hspace{-1em}(7,1),(7,8),(8,9),(9,10),(10,11),(11,12),(12,12)\}\\
0&\text{otherwise.}
\end{cases}
\end{eqnarray*}
The left-hand side of Figure \ref{fig:GM_abstraction} illustrates this result. The arrows are illustrations of the function $n$ that maps each realisation of $D^{t-1}$ to the corresponding realisation of $D^{t}$.
%%%
\begin{figure}[t]
\centering
\includegraphics[scale=0.5]{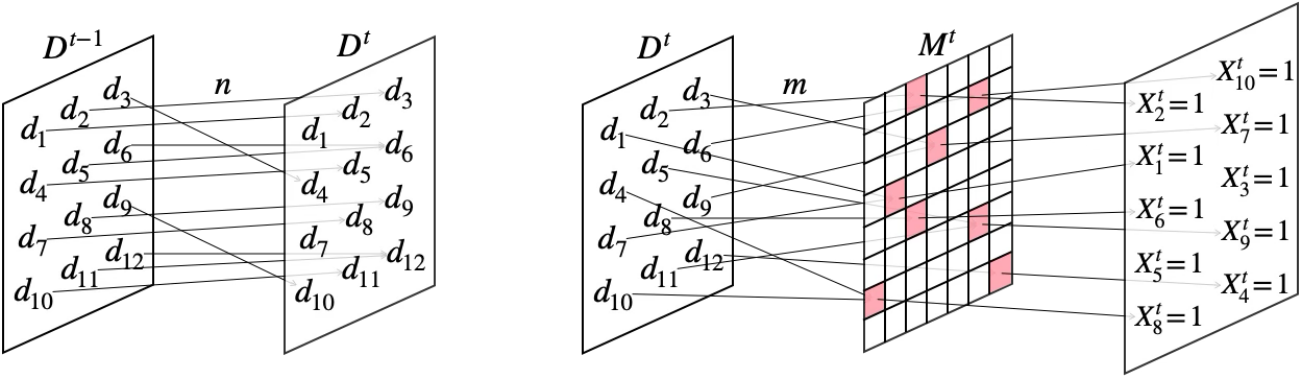}
\caption{Left: The data transition. Right: The support relations among data, models in propositional logic, and propositional formulas over time. There is an arrow from a data point to a model if the data point evidences the model, which illustrates the function $m$. There is an arrow from a model to a formula if the formula is true in the model.}
\label{fig:GM_abstraction}
\end{figure}
%%%
\qed
\end{example}
From Proposition \ref{prop:data}, Equation (\ref{eq:FJD}) can be simplified as follows.
\begin{eqnarray}
p(D^{1:T},M^{1:T},X_{1:I}^{1:T})&=&\prod_{t=1}^{T}\biggr[p(D^{t}|D^{t-1})p(M^{t}|D^{1:t},M^{1:t-1},X_{1:I}^{1:t-1})\biggr.\nonumber\\
&&\biggr.\prod_{i=1}^{I}p(X_{i}^{t}|D^{1:t},M^{1:t},X_{1:I}^{1:t-1},X_{1:i-1}^{t})\biggr]\label{eq:FJD2}
\end{eqnarray}
%%%%%%%%%%%%%%%%%%%%%%%%%%%%%%%%%%%%%%%%%%%%%%%%%%%%%%%%%%%%%%%%%%%%%%%%%%%%%%%%%%%%%%%%%%%%%%%%%%%%%%%%%%%%%%%%%%%%%%%%%%%%%%%%%%%%%%%%%%%%%%%%%%%%%%%%%%%%%%%%%%%%%%%%%%%%%%%%%%%%%%%%%%%%%%%%%%%%%%%%%%%%%%%%%%%%%%%%
\subsection{Model distributions}\label{sec:model_distribution}
In this section, we define the conditional probability of models, which appears in Equation (\ref{eq:FJD2}), and then analyse its property to further simply the equation. Each model in propositional logic is meant to represent a state of the world. It is thus natural to think that each model is supported or evidenced by data observed from the environment. We assume a function, $m: Data\to Models$, that maps each data point to the corresponding model supported by the data.
%%%
\begin{definition}\label{def:models}
For $t\in\{1,2,...,T\}$, the conditional probability of $m^{t}$ given $d^{1:t}$, $m^{1:t-1}$ and $x_{1:I}^{1:t-1}$ is defined as follows.
\begin{eqnarray*}
p(m^{t}|d^{1:t},m^{1:t-1},x_{1:I}^{1:t-1})=
\begin{cases}
1 & \text{if }m^{t}=m(d^{t})\\
0 & \text{otherwise}
\end{cases}
\end{eqnarray*}
\end{definition}
%%%
We derive the following property from Definition \ref{def:models}
\begin{proposition}\label{prop:models}
Let $t\in\{1,2,...,T\}$. $M^{t}$ is conditionally independent of $D^{1:t-1}$, $M^{1:t-1}$ and $X_{1:I}^{1:t-1}$ given $D^{t}$, i.e.,
\begin{eqnarray*}
p(M^{t}|D^{1:t},M^{1:t-1},X_{1:I}^{1:t-1})=p(M^{t}|D^{t}).
\end{eqnarray*}
\end{proposition}
%%%
\begin{proof}
The proof has the same structure as Proposition \ref{prop:data}.
\begin{eqnarray*}
&&p(m^{t}|d^{t})=\frac{p(d^{t},m^{t})}{p(d^{t})}=\frac{\sum_{d^{1:t-1}}\sum_{m^{1:t-1}}\sum_{x_{1:I}^{1:t-1}}p(d^{1:t},m^{1:t},x_{1:I}^{1:t-1})}{p(d^{t})}\\
&&=\frac{\sum_{d^{1:t-1}}\sum_{m^{1:t-1}}\sum_{x_{1:I}^{1:t-1}}p(m^{t}|d^{1:t},m^{1:t-1},x_{1:I}^{1:t-1})p(d^{1:t},m^{1:t-1},x_{1:I}^{1:t-1})}{p(d^{t})}
\end{eqnarray*}
By definition, the value of $p(m^{t}|d^{1:t},m^{1:t-1},x_{1:I}^{1:t-1})$ depends only on $d^{t}$. Therefore, the conditional probability can be moved to the outside of the summations.
\begin{eqnarray*}
&&=\frac{p(m^{t}|d^{1:t},m^{1:t-1},x_{1:I}^{1:t-1})\sum_{d^{1:t-1}}\sum_{m^{1:t-1}}\sum_{x_{1:I}^{1:t-1}}p(d^{1:t},m^{1:t-1},x_{1:I}^{1:t-1})}{p(d^{t})}\\
&&=\frac{p(m^{t}|d^{1:t},m^{1:t-1},x_{1:I}^{1:t-1})p(d^{t})}{p(d^{t})}=p(m^{t}|d^{1:t},m^{1:t-1},x_{1:I}^{1:t-1})
\end{eqnarray*}
Here, the second line is an application of the sum rule for marginalisation.
\qed
\end{proof}
%%%
From Proposition \ref{prop:models}, Equation (\ref{eq:FJD2}) can be simplified as follows.
\begin{eqnarray}
p(D^{1:T},M^{1:T},X_{1:I}^{1:T})&=&\prod_{t=1}^{T}\biggr[p(D^{t}|D^{t-1})p(M^{t}|D^{t})\biggr.\nonumber\\
&&\biggr.\prod_{i=1}^{I}p(X_{i}^{t}|D^{1:t},M^{1:t},X_{1:I}^{1:t-1},X_{1:i-1}^{t})\biggr]\label{eq:FJD3}
\end{eqnarray}
%%%%%%%%%%%%%%%%%%%%%%%%%%%%%%%%%%%%%%%%%%%%%%%%%%%%%%%%%%%%%%%%%%%%%%%%%%%%%%%%%%%%%%%%%%%%%%%%%%%%%%%%%%%%%%%%%%%%%%%%%%%%%%%%%%%%%%%%%%%%%%%%%%%%%%%%%%%%%%%%%%%%%%%%%%%%%%%%%%%%%%%%%%%%%%%%%%%%%%%%%%%%%%%%%%%%%%%%%%%%%%%%%%%%%%%%%%%%%%%%%%%%%%%%%%%%%%%%%%%%%%%%%%%%%%%%%%%%%%%%%%%%%%%%%%%%
\subsection{Knowledge distributions}\label{sec:knowledge_distribution}
In this section, we define the conditional probability of formulas, which appears in Equation (\ref{eq:FJD3}), and then analyse its property to further simply the equation. As usual, the truth value of a formula is determined solely in light of a model based on the semantics of propositional logic. We use the symbol $\ms{X_{i}}_{m_{l}}$ to denote the truth value of the formula $X_{i}\in{\cal L}$ in the model $m_{l}\in Models$.
\begin{definition}\label{def:formulas}
Let $\mu\in[0.5,1]$ and $t\in\{1,2,...,T\}$. The conditional probability of $x_{i}^{t}$ given $d^{1:t}$, $m^{1:t}$, $x_{1:I}^{1:t-1}$ and $x_{1:i-1}^{t}$ is defined as follows.
\begin{eqnarray*}
p(x_{i}^{t}|d^{1:t},m^{1:t},x_{1:I}^{1:t-1},x_{1:i-1}^{t})=
\begin{cases}
\mu & \text{if }x_{i}^{t}=\ms{X_{i}}_{m^{t}}\\
1-\mu & \text{otherwise}
\end{cases}
\end{eqnarray*}
\end{definition}
Namely, the truth value of a formula at a time point depends only on the model at the same time point. We derive the following property from Definition \ref{def:formulas}.
\begin{proposition}\label{prop:formulas}
Let $t\in\{1,2,...,T\}$ and $i\in\{1,2,...,I\}$. $X_{i}^{t}$ is conditionally independent of $D^{1:t}$, $M^{1:t-1}$, $X_{1:I}^{1:t-1}$ and $X_{1:i-1}^{t}$ given $M^{t}$, i.e.,
\begin{eqnarray*}
p(X_{i}^{t}|D^{1:t},M^{1:t},X_{1:I}^{1:t-1},X_{1:i-1}^{t})=p(X_{i}^{t}|M^{t}).
\end{eqnarray*}
\end{proposition}
\begin{proof}
The proof has the same structure as Propositions \ref{prop:data} and \ref{prop:models}.
{\small
\begin{eqnarray*}
&&p(x_{i}^{t}|m^{t})=\frac{p(m^{t},x_{i}^{t})}{p(m^{t})}=\frac{\sum_{d^{1:t}}\sum_{m^{1:t-1}}\sum_{x_{1:I}^{1:t-1}}\sum_{x_{1:i-1}^{t}}p(d^{1:t},m^{1:t},x_{1:I}^{1:t-1},x_{1:i}^{t})}{p(m^{t})}=\\
&&\frac{\sum_{d^{1:t}}\sum_{m^{1:t-1}}\sum_{x_{1:I}^{1:t-1}}\sum_{x_{1:i-1}^{t}}p(x_{i}^{t}|d^{1:t},m^{1:t},x_{1:I}^{1:t-1},x_{1:i-1}^{t})p(d^{1:t},m^{1:t},x_{1:I}^{1:t-1},x_{1:i-1}^{t})}{p(m^{t})}
\end{eqnarray*}
}
By definition, the value of $p(x_{i}^{t}|d^{1:t},m^{1:t},x_{1:I}^{1:t-1},x_{1:i-1}^{t})$ depends only on $m^{t}$. Therefore, the conditional probability can be moved to the outside of the summations.
{\small
\begin{eqnarray*}
&&=\frac{p(x_{i}^{t}|d^{1:t},m^{1:t},x_{1:I}^{1:t-1},x_{1:i-1}^{t})\sum_{d^{1:t}}\sum_{m^{1:t-1}}\sum_{x_{1:I}^{1:t-1}}\sum_{x_{1:i-1}^{t}}p(d^{1:t},m^{1:t},x_{1:I}^{1:t-1},x_{1:i-1}^{t})}{p(m^{t})}\\
&&=\frac{p(x_{i}^{t}|d^{1:t},m^{1:t},x_{1:I}^{1:t-1},x_{1:i-1}^{t})p(m^{t})}{p(m^{t})}=p(x_{i}^{t}|d^{1:t},m^{1:t},x_{1:I}^{1:t-1},x_{1:i-1}^{t})
\end{eqnarray*}
}
Here, the second line is an application of the sum rule for marginalisation.
\qed
\end{proof}
%%%
\begin{example}[Continued from Example \ref{ex:data}]\label{ex:formulas}
The data shown in Figure \ref{fig:maze} give rise to the following results regardless of the value of $t$.
\begin{eqnarray*}
&&p(X_{i}^{t}=1|M^{t}=m(d_{k}))=
\begin{cases}
\mu&\text{if } (k,i)\in\{(1,1),(2,2),(3,7),(4,8),(5,9),\\
&\hspace{-2.5em}(6,10),(7,1),(8,6),(9,7),(10,8),(11,9),(12,4)\}\\
1-\mu&\text{otherwise }
\end{cases}
\end{eqnarray*}
Here, recall that $m(d_{k})$ is the model supported by data $d_{k}$. The right-hand side of Figure \ref{fig:GM_abstraction} illustrates this result. Note that the hierarchy represents an abstraction relation in the sense that an element on each layer is selective ignorance of elements of its left layer. In fact, the truth value of each formula is determined once a model is given, but not vice versa. Each model is determined once a data point is given, but not vice versa.
\qed
\end{example}
%%%
From Proposition \ref{prop:formulas}, Equation (\ref{eq:FJD3}) can be simplified as follows.
\begin{theorem}\label{theo:FJD}
The full joint distribution over $D^{1:T}$, $M^{1:T}$ and $X_{1:I}^{1:T}$ is given as follows.
\begin{eqnarray}
p(D^{1:T},M^{1:T},X_{1:I}^{1:T})&=&\prod_{t=1}^{T}\biggr[p(D^{t}|D^{t-1})p(M^{t}|D^{t})\prod_{i=1}^{I}p(X_{i}^{t}|M^{t})\biggr]\label{eq:FJD4}
\end{eqnarray}
\end{theorem}
\begin{proof}
Applications of Propositions \ref{prop:data}, \ref{prop:models} and \ref{prop:formulas}.
\end{proof}
%%%
Equation (\ref{eq:FJD4}) is the simplest form of the full joint distribution. The centre graph of Figure \ref{fig:GMs} illustrates the equation, where there are arrows from each of the conditions to the outcome, for all the conditional probabilities appearing in the equation.
\par
In many cases, we are interested in reasoning over formulas. Obviously, the marginal distribution over formulas, i.e., Equation (\ref{eq:MJD}), can be written as follows using Equation (\ref{eq:FJD4}).
\begin{eqnarray}
p(X_{1:I}^{1:T})&=&\sum_{d^{1:T}\in Data^{T}}\sum_{m^{1:T}\in Models^{T}}\prod_{t=1}^{T}\biggr[p(d^{t}|d^{t-1})p(m^{t}|d^{t})\prod_{i=1}^{I}p(X_{i}^{t}|m^{t})\biggr]\label{eq:MJD2}
\end{eqnarray}
Interestingly, Equation (\ref{eq:MJD2}) can be further simplified. Let $n^{t}(d_{k})$ denote the data obtained by applying the function $n$ to the data $d_{k}$ $t$ times.
\begin{theorem}\label{theo:MJD}
The marginal distribution over $X_{1:I}^{1:T}$ is given as follows.
\begin{eqnarray}
p(X_{1:I}^{1:T})=\frac{1}{K}\sum_{k=1}^{K}\prod_{t=1}^{T}\prod_{i=1}^{I}p(X_{i}^{t}|m(n^{t-1}(d_{k})))\label{eq:MJD3}
\end{eqnarray}
\end{theorem}
\begin{proof}
Equation (\ref{eq:MJD2}) can be developed as follows by expanding the product over time and the summations over models.
\begin{eqnarray*}
&&\sum_{d^{1:T}}\sum_{m^{1:T}}\biggr[p(d^{1})p(m^{1}|d^{1})\prod_{i=1}^{I}p(X_{i}^{1}|m^{1})... p(d^{T}|d^{T-1})p(m^{T}|d^{T})\prod_{i=1}^{I}p(X_{i}^{T}|m^{T})\biggr]\\
&=&\sum_{d^{1:T}}\biggr[p(d^{1})\sum_{m^{1}}\biggr[p(m^{1}|d^{1})\prod_{i=1}^{I}p(X_{i}^{1}|m^{1})\biggr]... p(d^{T}|d^{T-1})\sum_{m^{T}}\biggr[p(m^{T}|d^{T})\prod_{i=1}^{I}p(X_{i}^{T}|m^{T})\biggr]\biggr]
\end{eqnarray*}
By definition, each data point supports a single model. We can thus remove the summations over models.
\begin{eqnarray*}
&=&\sum_{d^{1:T}}\biggr[p(d^{1})\prod_{i=1}^{I}p(X_{i}^{1}|m(d^{1}))... p(d^{T}|d^{T-1})\prod_{i=1}^{I}p(X_{i}^{T}|m(d^{T}))\biggr]
\end{eqnarray*}
Expanding the summations over data, we have 
\begin{eqnarray*}
&=&\sum_{d^{1}}\biggr[p(d^{1})\prod_{i=1}^{I}p(X_{i}^{1}|m(d^{1}))... \sum_{d^{T}}\biggr[p(d^{T}|d^{T-1})\prod_{i=1}^{I}p(X_{i}^{T}|m(d^{T}))\biggr]...\biggr].
\end{eqnarray*}
By definition, data changes deterministically. We can thus remove the summations over data, for all time steps except $t=1$.
\begin{eqnarray*}
&=&\sum_{d^{1}}\biggr[p(d^{1})\prod_{i=1}^{I}p(X_{i}^{1}|m(d^{1}))...\prod_{i=1}^{I}p(X_{i}^{T}|m(n^{T-1}(d^{1})))\biggr]
\end{eqnarray*}
Here, $n^{T-1}(d^{1})=n(d^{T-1})$. Since $p(D^{1})$ is the uniform distribution over the $K$ data, i.e., $Data=\{d_{1},d_{2},...,d_{K}\}$, we finally have
\begin{eqnarray*}
p(X_{1:I}^{1:T})&=&\frac{1}{K}\sum_{k=1}^{K}\prod_{t=1}^{T}\prod_{i=1}^{I}p(X_{i}^{t}|m(n^{t-1}(d_{k}))).
\end{eqnarray*}
\qed
\end{proof}
Theorem \ref{theo:MJD} shows the simplest form of the marginal distribution over formulas. The right-hand side of Figure \ref{fig:GMs} illustrates the result. There are arrows from the condition to the outcome, for all the conditional probabilities appearing in Equation (\ref{eq:MJD3}). Theorem \ref{theo:MJD} is computationally important as the omitted summation multiplication $\sum_{d^{2}}\sum_{d^{3}}\cdots\sum_{d^{T}}\sum_{m^{1}}\sum_{m^{2}}\cdots\sum_{m^{T}}$ does not change the result but is computationally intractable. For example, since $|Data|=12$ and $|Models|=2^{10}$ in Figure \ref{fig:maze}, Theorem \ref{theo:MJD} allows us to skip $(12\times 2^{10})^{T}$ steps.
\par
Let $\alpha$ and $\Delta$ be an element and a subset of $\{x_{i}^{t}|x_{i}^{t}\in x_{1:I}^{1:T}\}$, respectively. Using the sum rule and Theorem \ref{theo:MJD}, the conditional probability of $\alpha$ given $\Delta$ can be written as follows. 
\begin{eqnarray*}
p(\alpha|\Delta)&=&\frac{p(\alpha,\Delta)}{p(\Delta)}=\frac{\sum_{x_{1:I}^{1:T}(\notin\Delta\cup\{\alpha\})}p(x_{1:I}^{1:T})}{\sum_{x_{1:I}^{1:T}(\notin\Delta)}p(x_{1:I}^{1:T})}\\
&=&\frac{\sum_{k=1}^{K}\sum_{x_{1:I}^{1:T}(\notin\Delta\cup\{\alpha\})}\prod_{t=1}^{T}\prod_{i=1}^{I}p(x_{i}^{t}|m(n^{t-1}(d_{k})))}{\sum_{k=1}^{K}\sum_{x_{1:I}^{1:T}(\notin\Delta)}\prod_{t=1}^{T}\prod_{i=1}^{I}p(x_{i}^{t}|m(n^{t-1}(d_{k})))}
\end{eqnarray*}
For all $x_{i}^{t}\notin\Delta\cup\{\alpha\}$, $\sum_{x_{i}^{t}}p(x_{i}^{t}|m(n^{t-1}(d_{k})))=\mu+(1-\mu)=1$. Therefore, 
\begin{eqnarray}
&=&\frac{\sum_{k=1}^{K}\prod_{x_{i}^{t}\in\Delta\cup\{\alpha\}}p(x_{i}^{t}|m(n^{t-1}(d_{k})))}{\sum_{k=1}^{K}\prod_{x_{i}^{t}\in\Delta}p(x_{i}^{t}|m(n^{t-1}(d_{k})))}.\label{eq:MJD_condition}
\end{eqnarray}
\par
The following property regarding the negation connective is useful.
\begin{proposition}
Let $X_{i}\in{\cal L}$ and $t\in\{1,2,...,T\}$. $p(X_{i}^{t}=0)=p(\lnot X_{i}^{t}=1)$.
\end{proposition}
\begin{proof}
Since the interpretation of $X_{i}$ conforms to the semantics of propositional logic, $\ms{X_{i}}_{m(n^{t-1}(d_{k}))}=0$ iff $\ms{\lnot X_{i}}_{m(n^{t-1}(d_{k}))}=1$. From Theorem \ref{theo:MJD},
\begin{eqnarray*}
p(X_{i}^{t}=0)&=&\frac{1}{K}\sum_{k=1}^{K}p(X_{i}^{t}=0|m(n^{t-1}(d_{k})))=\frac{1}{K}\sum_{k=1}^{K}p(\lnot X_{i}^{t}=1|m(n^{t-1}(d_{k})))\\
&=&p(\lnot X_{i}^{t}=1).
\end{eqnarray*}
This holds regardless of the value of $\mu\in[0.5,1]$.
\qed
\end{proof}
In what follows, we write $X_{i}^{t}=0$ as $\lnot X_{i}^{t}=1$, and then abbreviate this as $x_{i}^{t}$.
%%%%%%%%%%%%%%%%%%%%%%%%%%%%%%%%%%%%%%%%%%%%%%%%%%%%%%%%%%%%%%%%%%%%%%%%%%%%%%%%%%%%%%%%%%%%%%%%%%%%%%%%%%%%%%%%%%%%%%%%%%%%%%%%%%%%%%%%%%%%%%%%%%%%%%%%%%%%%%%%%%%%%%%%%%%%%%%%%%%%%%%%%%%%%%%%%%%%%%%%%%%%%%%%%%%%%%%%%%%%%%%%%%%%%%%%%%%%%%%%%%%%%%%%%%%%%%%%%%%%%%%%%%%%%%%%%%%%%%%%%%%%%%%%%%%%
\subsection{Examples}\label{sec:examples}
This section discusses examples of the probabilistic model we defined and then simplified in the previous section. To explain the role of $\mu$ introduced in Definition \ref{def:formulas}, we consider the three situations: $\mu$ substituted by $1$, $\mu$ approaching 1, and $\mu$ strictly less than 1, i.e., $\mu=1$, $\mu\to1$, and $\mu\in[0.5,1)$, respectively.
%%%
\begin{example}[Continued from Example \ref{ex:formulas}]\label{ex:p(x10|x2)}
Let $\mu=1$. What is the probability that the robot is in Room 10 at Time 5 given that it is in Room 2 at Time 1? Using Equation (\ref{eq:MJD_condition}) and $Data=\{d_{1},d_{2},...,d_{12}\}$, we have
\begin{eqnarray*}
p(x_{10}^{5}|x_{2}^{1})&=&\frac{\sum_{k=1}^{12}p(x_{2}^{1}|m(d_{k}))p(x_{10}^{5}|m(n^{4}(d_{k})))}{\sum_{k=1}^{12}p(x_{2}^{1}|m(d_{k}))}\\
&=&\frac{\sum_{k\in\{2\}}\mu^{2}+\sum_{k\in \{3\text{-}6\}}\mu(1-\mu)+\sum_{k\in \{1,7\text{-}12\}}(1-\mu)^{2}}{\sum_{k\in\{2\}}\mu+\sum_{k\in \{1,3\text{-}12\}}(1-\mu)}\\
&=&\frac{\mu^{2}+4(1-\mu)^{2}+7(1-\mu)^{2}}{\mu+11(1-\mu)}=\frac{1}{1}.
\end{eqnarray*}
This result is natural because there exists data showing that the robot was in Room 10 four time steps after being in Room 2.
\qed
\end{example}
%%%
Note that $\mu=1$ is not necessarily the case in Definition \ref{def:formulas}.
%%%
\begin{example}[Continued from Example \ref{ex:formulas}]\label{ex:p(x10|x2,x3,x8)}
Suppose that the robot was in Rooms 2, 3 and 8 at Time 1, 2 and 3, respectively. What is the probability that the robot is in Room 10 at Time 5, i.e., $p(x_{10}^{5}|x_{2}^{1},x_{3}^{2},x_{8}^{3})$. Let $\mu=1$. Using Equation (\ref{eq:MJD_condition}), we have
\begin{eqnarray}
&&p(x_{10}^{5}|x_{2}^{1},x_{3}^{2},x_{8}^{3})\nonumber\\
&=&\frac{\sum_{k=1}^{12}p(x_{2}^{1}|m(d_{k}))p(x_{3}^{2}|m(n(d_{k})))p(x_{8}^{3}|m(n^{2}(d_{k})))p(x_{10}^{5}|m(n^{4}(d_{k})))}{\sum_{k=1}^{12}p(x_{2}^{1}|m(d_{k}))p(x_{3}^{2}|m(n(d_{k})))p(x_{8}^{3}|m(n^{2}(d_{k})))}\label{eq:p(x10|x2,x3,x8)}\\
&=&\frac{\sum_{k\in\{2\}}\mu^{3}(1-\mu)+\sum_{k\in\{3\text{-}6,8\}}\mu(1-\mu)^{3}+\sum_{k\in\{1,7,9\text{-}12\}}(1-\mu)^{4}}{\sum_{k\in\{2\}}\mu^{2}(1-\mu)+\sum_{k\in\{8\}}\mu(1-\mu)^{2}+\sum_{k\in\{1,3\text{-}7,9\text{-}12\}}(1-\mu)^{3}}\nonumber\\
&=&\frac{\mu^{3}(1-\mu)+5\mu(1-\mu)^{3}+6(1-\mu)^{4}}{\mu^{2}(1-\mu)+\mu(1-\mu)^{2}+10(1-\mu)^{3}}=\frac{0}{0}.\nonumber
\end{eqnarray}
In contrast to Example \ref{ex:p(x10|x2)}, the probability is not defined due to division by zero. This is because the twelve data indicate that the robot has never been in Rooms 2, 3 and 8 in this order. Now, let $\mu$ approaching 1, i.e., $\mu\to1$. We then have
\begin{eqnarray}\label{eq:muto1}
p(x_{10}^{5}|x_{2}^{1},x_{3}^{2},x_{8}^{3})&=&\lim_{\mu\to 1}\frac{\mu^{3}(1-\mu)+5\mu(1-\mu)^{3}+6(1-\mu)^{4}}{\mu^{2}(1-\mu)+\mu(1-\mu)^{2}+10(1-\mu)^{3}}\nonumber\\
&=&\lim_{\mu\to 1}\frac{\mu^{3}+5\mu(1-\mu)^{2}+6(1-\mu)^{3}}{\mu^{2}+\mu(1-\mu)+10(1-\mu)^{2}}=\frac{1}{1}.
\end{eqnarray}
%%%
\begin{figure}[t]
\centering
\begin{minipage}[c]{.5\textwidth}
\centering
\includegraphics[scale=0.22]{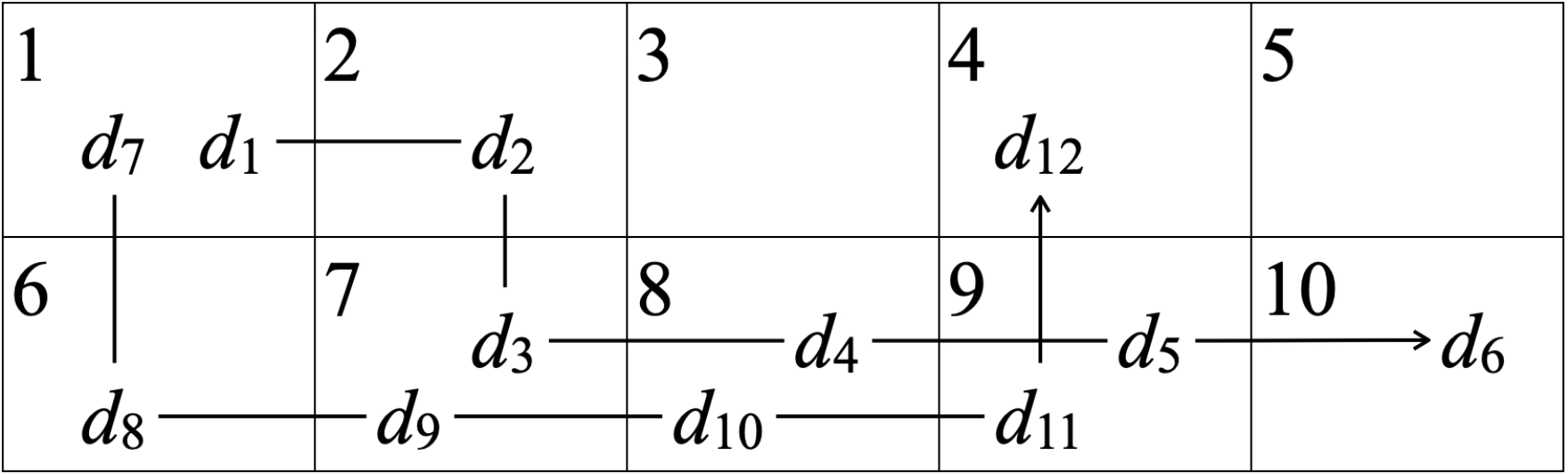}
\end{minipage}%
\begin{minipage}[c]{.5\textwidth}
\centering
\includegraphics[scale=0.22]{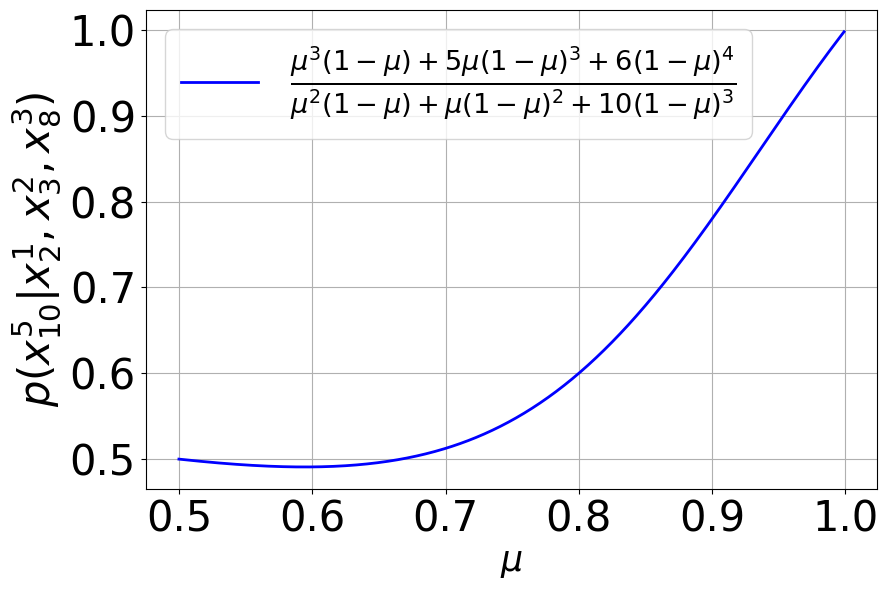}
\end{minipage}
\caption{Left: Figure \ref{fig:maze}. Right: $p(x_{10}^{5}|x_{2}^{1},x_{3}^{2},x_{8}^{3})$ as a function of $\mu$. The case of $\mu = 1$, which corresponds to the semantics of propositional logic, results in an undefined probability. This singularity can be resolved by taking the limit as $\mu \to 1$.}
\label{fig:plot}
\end{figure}
%%%
In Equation (\ref{eq:p(x10|x2,x3,x8)}), the summation in the denominator runs over all sequences of three consecutive data points (i.e., a sliding window of size 3), whereas the summation in the numerator runs over all sequences of five consecutive data points. In Equation (\ref{eq:muto1}), we can cancel $(1-\mu)$ that corresponds to the inconsistency between the condition $(x_{2}^{1},x_{3}^{2},x_{8}^{3})$ and the formulas satisfied by the best three consecutive data $(d_{2},d_{3},d_{4})$, where $x_{2}$ and $x_{8}$ are true in the models supported by $d_{2}$ and $d_{4}$, respectively. The numerator turns out to be the number of five consecutive data points in which $x_{2}$, $x_{8}$ and $x_{10}$ are true in the models supported by the first, third, and fifth data points, respectively. The right-hand side of Figure \ref{fig:plot} shows the probability as a function of $\mu$. The probability is undefined due to division by zero when $\mu$ is substituted by 1, whereas the limit resolves the singularity by assigning a reasonable value as $\mu$ approaches 1.
\qed
\end{example}
%%%
Note that $\mu\in[0.5,1)$ admits a chance that a formula is true in a model where it is actually false. However, this does not imply an opposition to the semantics of propositional logic. Rather, we use and extend the semantics to handle formulas concerning unfounded information, such as unknown and even false information, that cannot be made true in light of available data. The paper \cite{kido:24-1} provides the logical justification of this aspect in terms of paraconsistent logic \cite{Priest:02,Carnielli:07}. The paper \cite{kido:24-2} further provides statistical and probabilistic justifications in terms of maximum likelihood estimation and Bayesian networks.
%%%
\begin{example}[Continued from Example \ref{ex:formulas}]\label{ex:mu<1}
We show that $\mu\in[0.5,1)$ plays an important role that cannot be fulfilled when $\mu = 1$ or $\mu \to 1$. Consider the data shown on the left-hand side in Figure \ref{fig:maze_zero_frequency}. Using Equation (\ref{eq:p(x10|x2,x3,x8)}), we have
\begin{eqnarray*}
&&p(x_{10}^{5}|x_{2}^{1},x_{3}^{2},x_{8}^{3})=\frac{\overbrace{\mu^{3}(1-\mu)}^{d_{k}~s.t.~k\in\{2\}}+\overbrace{8\mu(1-\mu)^{3}}^{k\in\{3\text{-}7,9,12,14\}}+\overbrace{7(1-\mu)^{4}}^{k\in\{1,8,10,11,13,15,16\}}}{\underbrace{\mu^{2}(1-\mu)}_{k\in\{2\}}+\underbrace{4\mu(1-\mu)^{2}}_{k\in\{7,9,12,14\}}+\underbrace{11(1-\mu)^{3}}_{k\in\{1,3\text{-}6,8,10,11,13,15,16\}}}\\
&&p(x_{4}^{5}|x_{2}^{1},x_{3}^{2},x_{8}^{3})=\frac{\overbrace{5\mu^{2}(1-\mu)^{2}}^{d_{k}~s.t.~k\in\{2,7,9,12,14\}}+\overbrace{6\mu(1-\mu)^{3}}^{k\in\{8,10,11,13,15,16\}}+\overbrace{5(1-\mu)^{4}}^{k\in\{1,3\text{-}6\}}}{\underbrace{\mu^{2}(1-\mu)}_{d_{k}~s.t.~k\in\{2\}}+\underbrace{4\mu(1-\mu)^{2}}_{k\in\{7,9,12,14\}}+\underbrace{11(1-\mu)^{3}}_{k\in\{1,3\text{-}6,8,10,11,13,15,16\}}}.
\end{eqnarray*}
The right-hand side of Figure \ref{fig:maze_zero_frequency} shows these probabilities as functions of $\mu$. The three consecutive data starting from $d_{2}$ best match the condition, i.e., $x_{2}^{1}$, $x_{3}^{2}$ and $x_{8}^{3}$. Specifically, $d_{2}$ supports the model in which the robot is in Room 2 at Time 1, and the data two time steps later, $d_{4}$, supports the model in which the robot is in Room 8 at Time 3. This leads to the prediction that the robot will be in Room 10 at Time 5. $\mu\gtrsim 0.8$ reflects this fact.
\par
Meanwhile, the three consecutive data starting from $d_{7}$, $d_{9}$, $d_{12}$ and $d_{14}$ all next best match the same condition. Specifically, $d_{7}$ implies that $d_{9}$ supports the model in which the robot is in Room 8 at Time 3, and $d_{9}$ implies that $d_{10}$ supports the model in which the robot is in Room 3 at Time 2. These lead to the different prediction that the robot will be in Room 4 at Time 5. $\mu\lesssim 0.8$ reflects the situation where the matching quantity surpasses the matching quality.
%%%
\begin{figure}[t]
\centering
\begin{minipage}[c]{.5\textwidth}
\centering
\includegraphics[scale=0.22]{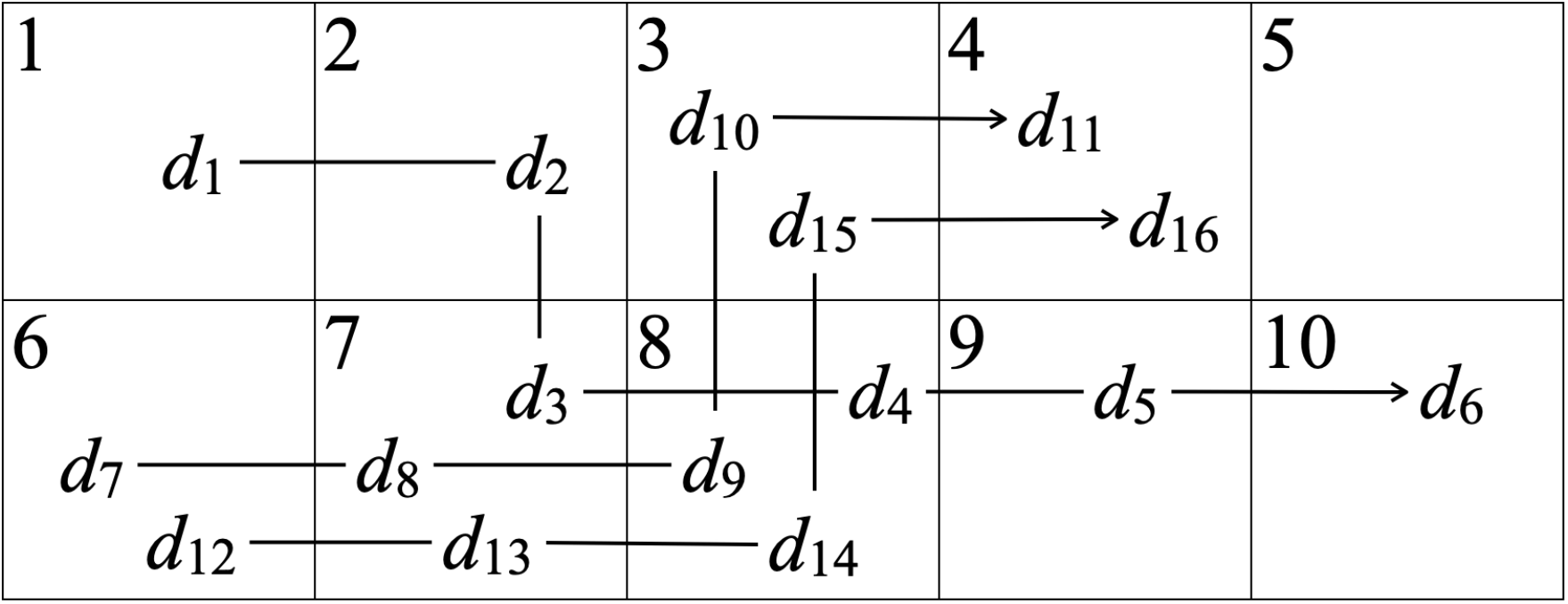}
\end{minipage}%
\begin{minipage}[c]{.5\textwidth}
\centering
\includegraphics[scale=0.22]{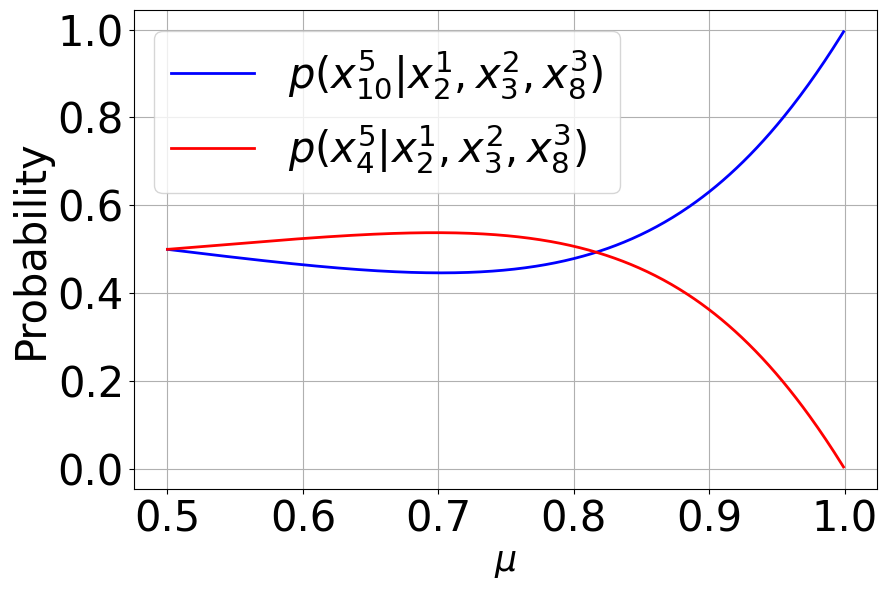}
\end{minipage}
\caption{Left: $Data=\{d_{1},d_{2},...,d_{16}\}$ collected in the same environment as shown in Figure \ref{fig:plot}. Right: The matching quantity is favoured over the matching quality when $\mu\lesssim0.8$.}
\label{fig:maze_zero_frequency}
\end{figure}
%%%
\qed
\end{example}
%%%%%%%%%%%%%%%%%%%%%%%%%%%%%%%%%%%%%%%%%%%%%%%%%%%%%%%%%%%%%%%%%%%%%%%%%%%%%%%%%%%%%%%%%%%%%%%%%%%%%%%%%%%%%%%%%%%%%%%%%%%%%%%%%%%%%%%%%%%%%%%%%%%%%%%%%%%%%%%%%%%%%%%%%%%%%%%%%%%%%%%%%%%%%%%%%%%%%%%%%%%%
\section{Evaluations}\label{sec:evaluation}
\subsection{Markov chains}\label{sec:MCs}
In this section, we compare the probabilistic model of abstraction with the $n$th-order, discrete-time, time-homogeneous Markov chains and hidden Markov models. Let $States=\{1,2,...,N\}$ be the set of natural numbers for $N$ states. For any discrete time $t\in\{1,2,...,T\}$, $S^{t}$ is a random variable taking values in $States$. $S^{t}=i$ represents that the state is $i$ at Time $t$. The $n$th-order Markov chain defines the full joint distribution as follows. 
\begin{eqnarray}\label{eq:nMC}
p(S^{1:T})=\prod_{t=1}^{T}p(S^{t}|S^{t-n:t-1})
\end{eqnarray}
Here, we ignore states with time zero or negative times. For example, $p(S^{2}|S^{-1:1})=p(S^{2}|S^{1})$. Maximum likelihood estimation is the statistical method most commonly used to estimate the parameters of probabilistic models solely from data. It is known that the maximum likelihood estimate for a categorical distribution is relative frequency \cite{Russell:20}. Equation (\ref{eq:nMC}) can then be written as follows.
\begin{eqnarray}\label{eq:MLEs}
p(S^{1:T})=\prod_{t=1}^{T}\frac{|S^{t-n:t}|}{|S^{t-n:t-1}|}
\end{eqnarray}
Here, $|S^{i:j}|$ denotes the number of sequences of consecutive data satisfying $S^{i:j}$. Let $\Delta\subseteq\{s^{t}|s^{t}\in s^{1:T}\}$. Using the sum rule and Equation (\ref{eq:MLEs}),
\begin{eqnarray}\label{eq:nMC_marginalisation}
&&p(\Delta)=\sum_{s^{1:T}(\notin\Delta)}p(s^{1:T})=\sum_{s^{1:T}(\notin\Delta)}\prod_{t=1}^{T}\frac{|s^{t-n:t}|}{|s^{t-n:t-1}|}\nonumber\\
&&=
\begin{cases}
\sum_{s^{1:T}(\notin\Delta)}\frac{\cancel{|s^{1}|}}{|()|}\frac{|s^{1:2}|}{\cancel{|s^{1}|}}\frac{|s^{2:3}|}{|s^{2}|}\frac{|s^{3:4}|}{|s^{3}|}\cdots\frac{|s^{T-2:T-1}|}{|s^{T-2}|}\frac{|s^{T-1:T}|}{|s^{T-1}|}&\text{if }n=1\\
\sum_{s^{1:T}(\notin\Delta)}\frac{\cancel{|s^{1}|}}{|()|}\frac{\cancel{|s^{1:2}|}}{\cancel{|s^{1}|}}\frac{|s^{1:3}|}{\cancel{|s^{1:2}|}}\frac{|s^{2:4}|}{|s^{2:3}|}\cdots\frac{|s^{T-3:T-1}|}{|s^{T-3:T-2}|}\frac{|s^{T-2:T}|}{|s^{T-2:T-1}|}&\text{if }n=2\\
\vdots\\
\sum_{s^{1:T}(\notin\Delta)}\frac{\cancel{|s^{1}|}}{|()|}\frac{\cancel{|s^{1:2}|}}{\cancel{|s^{1}|}}\frac{\cancel{|s^{1:3}|}}{\cancel{|s^{1:2}|}}\frac{\cancel{|s^{1:4}|}}{\cancel{|s^{1:3}|}}\cdots\frac{\cancel{|s^{1:T-1}|}}{\cancel{|s^{1:T-2}|}}\frac{|s^{1:T}|}{\cancel{|s^{1:T-1}|}}&\text{if }n=T-1.
\end{cases}
\end{eqnarray}
Here, $|()|$ denotes the number of data satisfying no constraints, and thus represents the total number of data.
%%%
\begin{figure}[t]
\begin{center}
\includegraphics[scale=0.45]{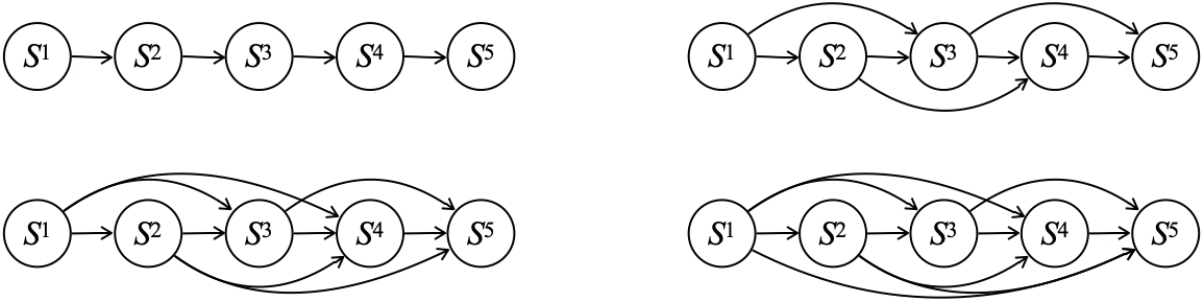}
\caption{Markov chains of order 1 (top left) to 4 (bottom right).}
\label{fig:nMCs}
\end{center}
\end{figure}
%%%
\begin{example}[Continued from Example \ref{ex:p(x10|x2)}]\label{ex:MCs}
Consider the same situation as in Example \ref{ex:p(x10|x2)}. $S^{t}=i$ represents that the robot is in Room $i$ at Time $t$, for all $i\in\{1,2,...,10\}$ and $t\in\{1,2,...,5\}$. We again ask the following question: What is the probability that the robot is in Room 10 at Time 5 given that it is in Room 2 at Time 1, i.e., $p(S^{5}=10|S^{1}=2)$? Figure \ref{fig:nMCs} shows the 1st-4th order Markov chains. Let $\#Z$ denote the cardinality of $Z$. Equation (\ref{eq:nMC_marginalisation}) can be expanded as follows.
\begin{eqnarray*}
&&p(S^{1}=2)=
\begin{cases}
\sum_{s^{2:5}}\frac{|S^{1}=2,s^{2}|}{|()|}\frac{|s^{2:3}|}{|s^{2}|}\frac{|s^{3:4}|}{|s^{3}|}\frac{|s^{4:5}|}{|s^{4}|}&\text{if }n=1\\
\sum_{s^{2:5}}\frac{|S^{1}=2,s^{2:3}|}{|()|}\frac{|s^{2:4}|}{|s^{2:3}|}\frac{|s^{3:5}|}{|s^{3:4}|}&\text{if }n=2\\
\sum_{s^{2:5}}\frac{|S^{1}=2,s^{2:4}|}{|()|}\frac{|s^{2:5}|}{|s^{2:4}|}&\text{if }n=3\\
\sum_{s^{2:5}}\frac{|S^{1}=2,s^{2:5}|}{|()|}&\text{if }n=4
\end{cases}\\
&&=
\begin{cases}
\frac{\#\{(d_{2},d_{3})\}}{\#Data}\frac{\#\{(d_{3},d_{4}),(d_{9},d_{10})\}}{\#\{(d_{3}),(d_{9})\}}\frac{\#\{(d_{4},d_{5}),(d_{10},d_{11})\}}{\#\{(d_{4}),(d_{10})\}}\frac{\#\{(d_{5},d_{6}),(d_{11},d_{12})\}}{\#\{(d_{5}),(d_{11})\}}=\frac{1}{12}\\
\frac{\#\{(d_{2},d_{3},d_{4})\}}{\#Data}\frac{\#\{(d_{3},d_{4},d_{5}),(d_{9},d_{10},d_{11})\}}{\#\{(d_{3},d_{4}),(d_{9},d_{10})\}}\frac{\#\{(d_{4},d_{5},d_{6}),(d_{10},d_{11},d_{12})\}}{\#\{(d_{4},d_{5}),(d_{10},d_{11})\}}=\frac{1}{12}\\
\frac{\#\{(d_{2},d_{3},d_{4},d_{5})\}}{\#Data}\frac{\#\{(d_{3},d_{4},d_{5},d_{6}),(d_{3},d_{4},d_{5},d_{12})\}}{\#\{(d_{3},d_{4},d_{5}),(d_{9},d_{10},d_{11})\}}=\frac{1}{12}\\
\frac{\#\{(d_{2},d_{3},d_{4},d_{5},d_{6})\}}{\#Data}=\frac{1}{12}
\end{cases}\\
&&p(S^{1}=2,S^{5}=10)=
\begin{cases}
\sum_{s^{2:4}}\frac{|S^{1}=2,s^{2}|}{|()|}\frac{|s^{2:3}|}{|s^{2}|}\frac{|s^{3:4}|}{|s^{3}|}\frac{|s^{4},S^{5}=10|}{|s^{4}|}&\text{if }n=1\\
\sum_{s^{2:4}}\frac{|S^{1}=2,s^{2:3}|}{|()|}\frac{|s^{2:4}|}{|s^{2:3}|}\frac{|s^{3:4},S^{5}=10|}{|s^{3:4}|}&\text{if }n=2\\
\sum_{s^{2:4}}\frac{|S^{1}=2,s^{2:4}|}{|()|}\frac{|s^{2:4},S^{5}=10|}{|s^{2:4}|}&\text{if }n=3\\
\sum_{s^{2:4}}\frac{|S^{1}=2,s^{2:4},S^{5}=10|}{|()|}&\text{if }n=4
\end{cases}\\
&&=
\begin{cases}
\frac{\#\{(d_{2},d_{3})\}}{\#Data}\frac{\#\{(d_{3},d_{4}),(d_{9},d_{10})\}}{\#\{(d_{3}),(d_{9})\}}\frac{\#\{(d_{4},d_{5}),(d_{10},d_{11})\}}{\#\{(d_{4}),(d_{10})\}}\frac{\#\{(d_{5},d_{6})\}}{\#\{(d_{5}),(d_{11})\}}=\frac{1}{24}\\
\frac{\#\{(d_{2},d_{3},d_{4})\}}{\#Data}\frac{\#\{(d_{3},d_{4},d_{5}),(d_{9},d_{10},d_{11})\}}{\#\{(d_{3},d_{4}),(d_{9},d_{10})\}}\frac{\#\{(d_{4},d_{5},d_{6})\}}{\#\{(d_{4},d_{5}),(d_{10},d_{11})\}}=\frac{1}{24}\\
\frac{\#\{(d_{2},d_{3},d_{4},d_{5})\}}{\#Data}\frac{\#\{(d_{3},d_{4},d_{5},d_{6})\}}{\#\{(d_{3},d_{4},d_{5}),(d_{9},d_{10},d_{11})\}}=\frac{1}{24}\\
\frac{\#\{(d_{2},d_{3},d_{4},d_{5},d_{6})\}}{\#Data}=\frac{1}{12}
\end{cases}\\
&&p(S^{5}=10|S^{1}=2)=\frac{p(S^{1}=2,S^{5}=10)}{p(S^{1}=2)}=
\begin{cases}
\frac{1/24}{1/12}=\frac{1}{2}&\text{if }n=1\\
\frac{1/24}{1/12}=\frac{1}{2}&\text{if }n=2\\
\frac{1/24}{1/12}=\frac{1}{2}&\text{if }n=3\\
\frac{1/12}{1/12}=1&\text{if }n=4
\end{cases}
\end{eqnarray*}
The left-hand side of Figure \ref{fig:plot} illustrates that the robot's presence in Room 2 is indicated only by $d_{2}$, and the data from four time steps later, i.e., $d_{6}$, shows that the robot is in Room 10. The above result shows that this analysis can be explained only by the 4th-order Markov chain. Indeed, the $n$th-order Markov chain can make a prediction based solely on the past $n$ states. 
\qed
\end{example}
%%%
Higher-order Markov chains offer greater flexibility in fitting data compared to lower-order ones. However, this flexibility comes at the expense of a large hypothesis space. Indeed, the $n$th-order Markov chains with $r$ states has $(r-1)\times r^{n}$ parameters, which leads to issues of computational intractability and data scarcity, e.g., zero frequency. For example, only $12$ parameters out of $90,000(=9\times 10^{4})$ in the 4th-order Markov chain can be learnt from the 12 data. Data smoothing is clearly not a fundamental solution to this issue.
\begin{example}[Continued from Example \ref{ex:p(x10|x2,x3,x8)}]\label{ex:zero}
We revisit Example \ref{ex:p(x10|x2,x3,x8)} and apply the Markov chains with parameters trained using maximum likelihood estimation. The following results can be obtained using Equation (\ref{eq:nMC_marginalisation}).
\begin{eqnarray*}
&&p(S^{5}=10|S^{1}=2,S^{2}=3,S^{3}=8)=\frac{p(S^{1}=2,S^{2}=3,S^{3}=8,S^{5}=10)}{p(S^{1}=2,S^{2}=3,S^{3}=8)}\\
&&=\begin{cases}
\frac{\sum_{s^{4}}\frac{|S^{1}=2,S^{2}=3|}{|()|}\frac{|S^{2}=3,S^{3}=8|}{|S^{2}=3|}\frac{|S^{3}=8,s^{4}|}{|S^{3}=8|}\frac{|s^{4},S^{5}=10|}{|s^{4}|}}{\sum_{s^{4:5}}\frac{|S^{1}=2,S^{2}=3|}{|()|}\frac{|S^{2}=3,S^{3}=8|}{|S^{2}=3|}\frac{|S^{3}=8,s^{4}|}{|S^{3}=8|}\frac{|s^{4:5}|}{|s^{4}|}}=\frac{\frac{0}{12}\frac{0}{0}\frac{2}{2}\frac{1}{2}}{\frac{0}{12}\frac{0}{0}\frac{2}{2}\frac{2}{2}}=\frac{0}{0}&\text{if }n=1\\
\frac{\sum_{s^{4}}\frac{|S^{1}=2,S^{2}=3,S^{3}=8|}{|()|}\frac{|S^{2}=3,S^{3}=8,s^{4}|}{|S^{2}=3,S^{3}=8|}\frac{|S^{3}=8,s^{4},S^{5}=10|}{|S^{3}=8,s^{4}|}}{\sum_{s^{4:5}}\frac{|S^{1}=2,S^{2}=3,S^{3}=8|}{|()|}\frac{|S^{2}=3,S^{3}=8,s^{4}|}{|S^{2}=3,S^{3}=8|}\frac{|S^{3}=8,s^{4:5}|}{|S^{3}=8,s^{4}|}}=\frac{\frac{0}{12}\frac{0}{0}\frac{1}{2}}{\frac{0}{12}\frac{0}{0}\frac{2}{2}}=\frac{0}{0}&\text{if }n=2\\
\frac{\sum_{s^{4}}\frac{|S^{1}=2,S^{2}=3,S^{3}=8,s^{4}|}{|()|}\frac{|S^{2}=3,S^{3}=8,s^{4},S^{5}=10|}{|S^{2}=3,S^{3}=8,s^{4}|}}{\sum_{s^{4:5}}\frac{|S^{1}=2,S^{2}=3,S^{3}=8,s^{4}|}{|()|}\frac{|S^{2}=3,S^{3}=8,s^{4:5}|}{|S^{2}=3,S^{3}=8,s^{4}|}}=\frac{\frac{0}{12}\frac{0}{0}}{\frac{0}{12}\frac{0}{0}}=\frac{0}{0}&\text{if }n=3\\
\frac{\sum_{s^{4}}\frac{|S^{1}=2,S^{2}=3,S^{3}=8,s^{4},S^{5}=10|}{|()|}}{\sum_{s^{4:5}}\frac{|S^{1}=2,S^{2}=3,S^{3}=8,s^{4:5}|}{|()|}}=\frac{\frac{0}{12}}{\frac{0}{12}}=\frac{0}{0}&\text{if }n=4
\end{cases}
\end{eqnarray*}
None of them is defined due to division by zero, meaning that no meaningful predictions are possible using the Markov chains regardless of their orders. Intuitively, the predictions fail since the robot did not experience Room $3$.
\qed
\end{example}
%%%%%%%%%%%%%%%%%%%%%%%%%%%%%%%%%%%%%%%%%%%%%%%%%%%%%%%%%%%%%%%%%%%%%%%%%%%%%%%%%%%%%%%%%%%%%%%%%%%%%%%%%%%%%%%%%%%%%%%%%%%%%%%%%%%%%%%%%%%%%%%%%%%%%%%%%%%%%%%%%%
\subsection{Equivalence}\label{sec:equivalence}
We discuss the relationship between Markov chains and the probabilistic model of abstraction. To relate the propositional language to Markov chains, we use the propositional variable $S_{i}^{t}$, which denotes that state $S$ has value $i$ at time $t$ in the Markov chain, i.e., $s_{i}^{t}$ (or $S_{i}^{t}=1$) iff $S^{t}=i$.
\par
For random variables or their realisations $z$, we assume that $p(z; n)$ and $p(z; \mu)$ represent the probability $p(z)$ obtained with an $n$th-order Markov model and with our probabilistic model with $\mu$, respectively. The symbol `;' denotes that its right-hand side is a variable, but not a random variable. We can show that our probabilistic model with $\mu=1$ and $\mu\to1$ and the highest-order, i.e., full-memory, Markov model trained using maximum likelihood estimation give the same joint distribution.
\begin{theorem}\label{theo:comp}
The following relation holds.
\begin{eqnarray*}
p(s_{h}^{1},s_{i}^{2},...,s_{j}^{T}; \mu=1)&=&p(s_{h}^{1},s_{i}^{2},...,s_{j}^{T}; \mu\to1)\\
&=&p(S^{1}=h,S^{2}=i,...,S^{T}=j; n=T-1)
\end{eqnarray*}
\end{theorem}
%%%
\begin{proof}
By definition, if formula $S_{i}$ is true in model $m^{t}$, i.e., $\ms{S_{i}}_{m^{t}}=1$, then $p(s_{i}^{t}|m^{t})=\mu=1$, for $\mu=1$ and $\mu\to1$. If $S_{i}$ is false in $m^{t}$, i.e., $\ms{S_{i}}_{m^{t}}=0$, then $p(s_{i}^{t}|m^{t})=1-\mu=0$, for $\mu=1$ and $\mu\to1$. We thus have
\begin{eqnarray*}
p(s_{h}^{1},s_{i}^{2},...,s_{j}^{T})&=&\frac{1}{K}\sum_{k=1}^{K}\biggr[p(s_{h}^{1}|m(d_{k}))p(s_{i}^{2}|m(n(d_{k})))\cdots p(s_{j}^{T}|m(n^{T-1}(d_{k})))\biggr]\\
&=&\frac{1}{K}\sum_{k=1}^{K}\biggr[\ms{S_{h}}_{m(d_{k})}\ms{S_{i}}_{m(n(d_{k}))}\cdots\ms{S_{j}}_{m(n^{T-1}(d_{k}))}\biggr].
\end{eqnarray*}
The expression inside the summation turns out to be one if $S_{h}$, $S_{i}$, ..., $S_{j}$ are sequentially true in the models supported by the $T$ consecutive data from $d_{k}$, and zero otherwise. Since $d_{k}$ ranges from $d_{1}$ to $d_{K}$, the summation turns out to be the number of such sequences. From Equation (\ref{eq:nMC_marginalisation}), we thus have
\begin{eqnarray*}
&=&\frac{|S^{1}=h, S^{2}=i,..., S^{T}=j|}{K}=p(S^{1}=h, S^{2}=i,..., S^{T}=j).
\end{eqnarray*}
\qed
\end{proof}
%%%
Any marginal distributions and conditional distributions can be derived from the joint distribution using valid rules of probability theory. Theorem \ref{theo:comp} thus establishes the equivalence between our probabilistic model and the highest-order Markov chain trained via maximum likelihood estimation.
%%%%%%%%%%%%%%%%%%%%%%%%%%%%%%%%%%%%%%%%%%%%%%%%%%%%%%%%%%%%%%%%%%%%%%%%%%%%%%%%%%%%%%%%%%%%%%%%%%%%%%%%%%%%%%%%%%%%%%%%%%%%%%%%%%%%%%%%%%%%%%%%%%%%%%%%%%%%%%%%%%
\subsection{Transparency}\label{sec:transparency}
\begin{figure}[t]
\centering
\begin{minipage}[c]{.5\textwidth}
\centering
\includegraphics[scale=0.25]{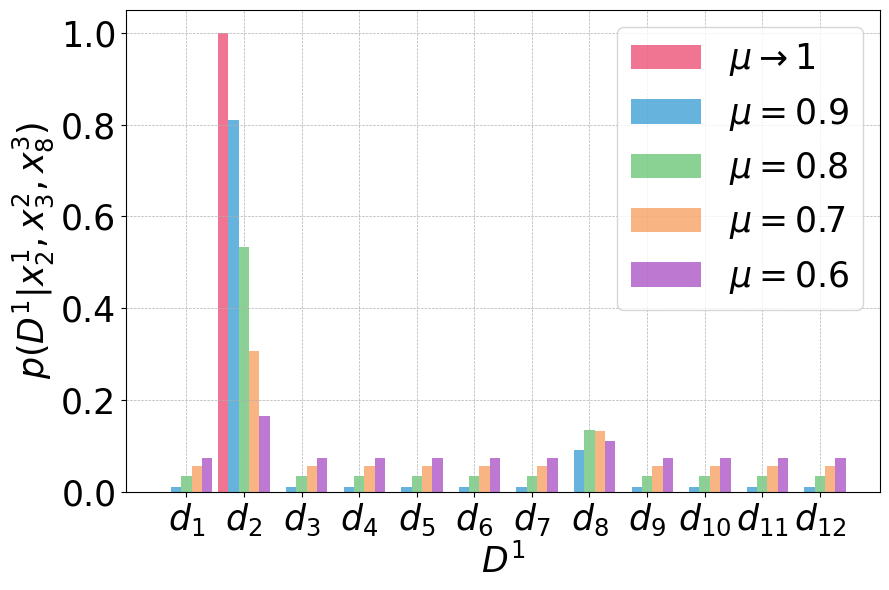}
\end{minipage}%
\begin{minipage}[c]{.5\textwidth}
\centering
\includegraphics[scale=0.25]{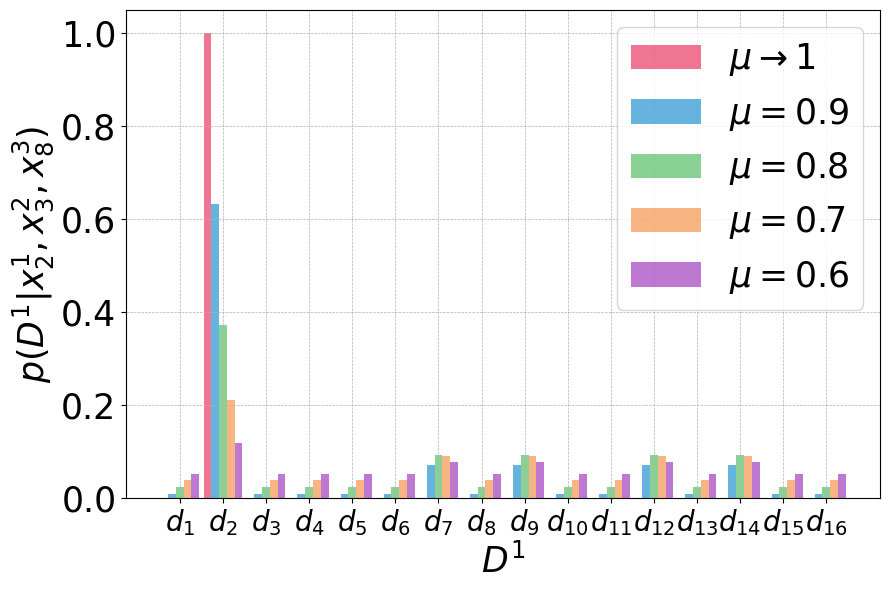}
\end{minipage}
\caption{The conditional data distributions make the data reference transparent. Left: $Data=\{d_{1},d_{2},...,d_{12}\}$ from Figure \ref{fig:plot}. Right: $Data=\{d_{1},d_{2},...,d_{16}\}$ from Figure \ref{fig:maze_zero_frequency}.}
\label{fig:data_transparency}
\end{figure}

Probabilistic modelling, including Markov chains, generally exhibits higher transparency compared to other modern machine learning paradigms such as deep learning and reinforcement learning. This is mainly because random variables and their dependencies are made explicit in probabilistic models. However, probabilistic modelling is not highly transparent from the data perspective. This is because reasoning operates using parameters rather than data. Indeed, learning is the process to exploit data to adjust the parameters of probabilistic models, whereas reasoning is the process to exploit the parameters, not the data itself, to make predictions. The following proposition states that the probabilistic model of abstraction over formulas always refers to data.
\begin{proposition}\label{prop:topdown}
Let $\alpha\in\{x_{i}^{t}|x_{i}^{t}\in x_{1:I}^{1:T}\}$ and $\Delta\subseteq\{x_{i}^{t}|x_{i}^{t}\in x_{1:I}^{1:T}\}$. The following relation holds.
\begin{eqnarray*}
p(\alpha|\Delta)=\sum_{d_{k}}p(\alpha|D^{1}=d_{k})p(D^{1}=d_{k}|\Delta)
\end{eqnarray*}
\end{proposition}
\begin{proof}
The left-hand side can be expanded as follows.
\begin{eqnarray*}
p(\alpha|\Delta)=\frac{p(\alpha,\Delta)}{p(\Delta)}=\frac{\sum_{d_{k}}p(\alpha,\Delta,D^{1}=d_{k})}{p(\Delta)}
\end{eqnarray*}
From the the rightmost graph of Figure \ref{fig:GMs}, we have
\begin{eqnarray*}
=\frac{\sum_{d_{k}}p(\alpha|d_{k})p(\Delta|d_{k})p(d_{k})}{p(\Delta)}=\frac{\sum_{d_{k}}p(\alpha|d_{k})p(d_{k}|\Delta)p(\Delta)}{p(\Delta)}=\sum_{d_{k}}p(\alpha|d_{k})p(d_{k}|\Delta).
\end{eqnarray*}
\qed
\end{proof}
Proposition \ref{prop:topdown} implies that reasoning over formulas is a sort of Bayesian learning \cite{Russell:20}. In Proposition \ref{prop:topdown}, data are marginalised out to infer formulas from given formulas. This fits the data-driven perspective. In typical probabilistic models, parameters are marginalised out to infer data from given data. This fits the model-driven perspective.
\begin{example}[Continued from Examples \ref{ex:p(x10|x2,x3,x8)} and \ref{ex:mu<1}]\label{ex:data_distribution}
Consider the left-hand side in Figure \ref{fig:plot} where $Data=\{d_{1},d_{2},...,d_{12}\}$. By definition, $d^{2:T}$ and $m^{1:T}$ are fully determined given $d^{1}$. The summations over these values thus can be omitted using $n^{t}(d^{1})$ and $m(n^{t}(d^{1}))$, for $t\in\{1,2,...,T-1\}$. Let $d_{k}\in Data$. Similar to Equation (\ref{eq:muto1}), we have
\begin{eqnarray*}
&&p(D^{1}=d_{k}|x_{2}^{1},x_{3}^{2},x_{8}^{3})=\frac{p(D^{1}=d_{k},x_{2}^{1},x_{3}^{2},x_{8}^{3})}{p(x_{2}^{1},x_{3}^{2},x_{8}^{3})}\\
&&=\frac{p(x_{2}^{1}|m(d_{k}))p(x_{3}^{2}|m(n(d_{k})))p(x_{8}^{3}|m(n^{2}(d_{k})))}{\sum_{k\in\{2\}}\mu^{2}(1-\mu)+\sum_{k\in\{8\}}\mu(1-\mu)^{2}+\sum_{k\in\{1,3\text{-}7,9\text{-}12\}}(1-\mu)^{3}}\\
&&=
\begin{cases}
\frac{\mu^{2}(1-\mu)^{2}}{\mu^{2}(1-\mu)+\mu(1-\mu)^{2}+10(1-\mu)^{3}}&\text{if }k\in\{2\}\\
\frac{\mu(1-\mu)^{2}}{\mu^{2}(1-\mu)+\mu(1-\mu)^{2}+10(1-\mu)^{3}}&\text{if }k\in\{8\}\\
\frac{(1-\mu)^{3}}{\mu^{2}(1-\mu)+\mu(1-\mu)^{2}+10(1-\mu)^{3}}&\text{if }k\in\{1,3\text{-}7,9\text{-}12\}.
\end{cases}
\end{eqnarray*}
The left-hand side of Figure \ref{fig:data_transparency} shows the conditional distribution over $Data=\{d_{1},d_{2},...,d_{12}\}$. The right-hand side shows the same distribution over $Data=\{d_{1},d_{2},...,d_{16}\}$ we discussed in Example \ref{ex:mu<1}.
\qed
\end{example}
%%%
%%%
%%%%%%%%%%%%%%%%%%%%%%%%%%%%%%%%%%%%%%%%%%%%%%%%%%%%%%%%%%%%%%%%%%%%%%%%%%%%%%%%%%%%%%%%%%%%%%%%%%%%%%%%%%%%%%%%%%%%%%%%%%%%%%%%%%%%%%%%%%%%%%%%%%%%%%%%%%%%%%%%%%
\subsection{Hidden states}\label{sec:hidden_states}
In Markov chains, it is typically assumed that the states of interest, often referred to as latent or hidden variables, are observable from the environment. The assumption does not hold in hidden Markov models, which instead assume that only effects, often referred to as observable variables, caused by these states are observable. The hidden and observable variables are clearly distinguished in the graphical models of hidden Markov models. In this section, we show that such a distinction is unnecessary for the probabilistic model of abstraction.
\begin{example}[Continued from Example \ref{ex:p(x10|x2)}]
Let us revisit Example \ref{ex:p(x10|x2)} and assume that the robot location is a hidden variable. Namely, the robot cannot detect its location from the environment using its own sensors. Instead, the robot is assumed to be able to perceive the presence of an obstacle in each direction. Let $N$, $E$, $S$, and $W$ be random variables representing the presence (denoted, e.g., by $N=1$ or $n$) and the absence (denoted, e.g., by $N=0$ or $\lnot n$) of an obstacle to north, east, south, and west, respectively. The graphical model of the probabilistic model of abstraction we need to handle this problem is depicted on the left-hand side in Figure \ref{fig:hidden_states}, where there is no structural distinction between the hidden and observable variables, i.e., the locations of robot and the presence of obstacles, respectively.
\par
Now, suppose that the robot was in Rooms $2$, $3$, and $8$ at Time $1$, $2$, and $3$, respectively. By the assumption, the robot only perceived $n$, $\lnot e$, $\lnot s$, and $\lnot w$ in Room $2$, $n$, $\lnot e$, $\lnot s$, and $\lnot w$ in Room $3$, and $\lnot n$, $\lnot e$, $s$, and $\lnot w$ in Room $8$. The probability of the robot being in Room $10$ at Time $5$ is given as follows.
\begin{eqnarray*}
&&p(x_{10}^{5}|n^{1},\lnot e^{1},\lnot s^{1},\lnot w^{1},n^{2},\lnot e^{2},\lnot s^{2},\lnot w^{2},\lnot n^{3},\lnot e^{3},s^{3},\lnot w^{3})=\\
&&\frac{\sum_{k=1}^{12}p(x_{10}^{5}|m(n^{4}(d_{k})))\prod_{a\in\{n,\lnot e,\lnot s,\lnot w\}}A\prod_{b\in\{n,\lnot e,\lnot s,\lnot w\}}B\prod_{c\in\{\lnot n,\lnot e, s,\lnot w\}}C}{\sum_{k=1}^{12}\prod_{a\in\{n,\lnot e,\lnot s,\lnot w\}}A\prod_{b\in\{n,\lnot e,\lnot s,\lnot w\}}B\prod_{c\in\{\lnot n,\lnot e, s,\lnot w\}}C}
\end{eqnarray*}
where $A=p(a^{1}|m(d_{k}))$, $B=p(b^{2}|m(n(d_{k})))$ and $C=p(c^{3}|m(n^{2}(d_{k})))$. Now,
\begin{eqnarray*}
&&\ms{n,\lnot e,\lnot s, w}_{m(d_{k})}=1\text{, for all }k\in\{1\}\\
&&\ms{n,\lnot e,\lnot s, \lnot w}_{m(d_{k})}=1\text{, for all }k\in\{2,12\}\\
&&\ms{\lnot n,\lnot e, s, \lnot w}_{m(d_{k})}=1\text{, for all }k\in\{3,4,5,9,10,11\}\\
&&\ms{\lnot n, e, s, \lnot w}_{m(d_{k})}=1\text{, for all }k\in\{6\}\\
&&\ms{n, \lnot e, \lnot s, w}_{m(d_{k})}=1\text{, for all }k\in\{7\}\\
&&\ms{\lnot n, \lnot e, s, w}_{m(d_{k})}=1\text{, for all }k\in\{8\}.
\end{eqnarray*}
Using Equation (\ref{eq:MJD_condition}), we thus have
{\small
\begin{eqnarray*}
&=&\frac{\overbrace{2\mu^{11}(1-\mu)^{2}}^{d_{1},d_{2}}+\overbrace{\mu^{10}(1-\mu)^{3}}^{d_{12}}+\overbrace{\mu^{9}(1-\mu)^{4}}^{d_{3}}+\overbrace{4\mu^{8}(1-\mu)^{5}}^{d_{4},d_{7},d_{9},d_{11}}+\overbrace{2\mu^{7}(1-\mu)^{6}}^{d_{5},d_{8}}+\overbrace{2\mu^{6}(1-\mu)^{7}}^{d_{6},d_{10}}}{\underbrace{\mu^{11}(1-\mu)}_{d_{1}}+\underbrace{2\mu^{10}(1-\mu)^{2}}_{d_{2},d_{12}}+\underbrace{4\mu^{8}(1-\mu)^{4}}_{d_{3},d_{7},d_{9},d_{11}}+\underbrace{2\mu^{7}(1-\mu)^{5}}_{d_{4},d_{8}}+\underbrace{2\mu^{6}(1-\mu)^{6}}_{d_{5},d_{10}}+\underbrace{\mu^{5}(1-\mu)^{7}}_{d_{6}}}.
\end{eqnarray*}
}
$\mu=1$ results in undefined values, whereas $\mu\in[0.5,1)$ allows us to cancel $\mu^{5}(1-\mu)$. In particular, as $\mu\to1$, we have
{\small
\begin{eqnarray*}
&&=\lim_{\mu\to1}\frac{2\mu^{6}(1-\mu)+\mu^{5}(1-\mu)^{2}+\mu^{4}(1-\mu)^{3}+4\mu^{3}(1-\mu)^{4}+2\mu^{2}(1-\mu)^{5}+2\mu(1-\mu)^{6}}{\mu^{6}+2\mu^{5}(1-\mu)+4\mu^{3}(1-\mu)^{3}+2\mu^{2}(1-\mu)^{4}+2\mu(1-\mu)^{5}+(1-\mu)^{6}}\\
&&=0.
\end{eqnarray*}
}
The right-hand side of Figure \ref{fig:hidden_states} illustrates this result, along with the probability of the robot being in Room 9 at Time 5. Similar to Example \ref{ex:data_distribution}, the data distribution explains why this is the case.
{\small
\begin{eqnarray*}
&&p(D^{1}=d_{1}|n^{1},\lnot e^{1},\lnot s^{1},\lnot w^{1},n^{2},\lnot e^{2},\lnot s^{2},\lnot w^{2},\lnot n^{3},\lnot e^{3},s^{3},\lnot w^{3})\\
&&=\lim_{\mu\to1}\frac{\mu^{6}}{\mu^{6}+2\mu^{5}(1-\mu)+4\mu^{3}(1-\mu)^{3}+2\mu^{2}(1-\mu)^{4}+2\mu(1-\mu)^{5}+(1-\mu)^{6}}=1
\end{eqnarray*}
}
Namely, only the three consecutive data from $d_{1}$ best explain the given presence of obstacles. This leads to the prediction that the robot location at Time $5$ is Room $9$, not Room $10$.
\qed
%%%
\begin{figure}[t]
\begin{center}
\includegraphics[scale=0.19]{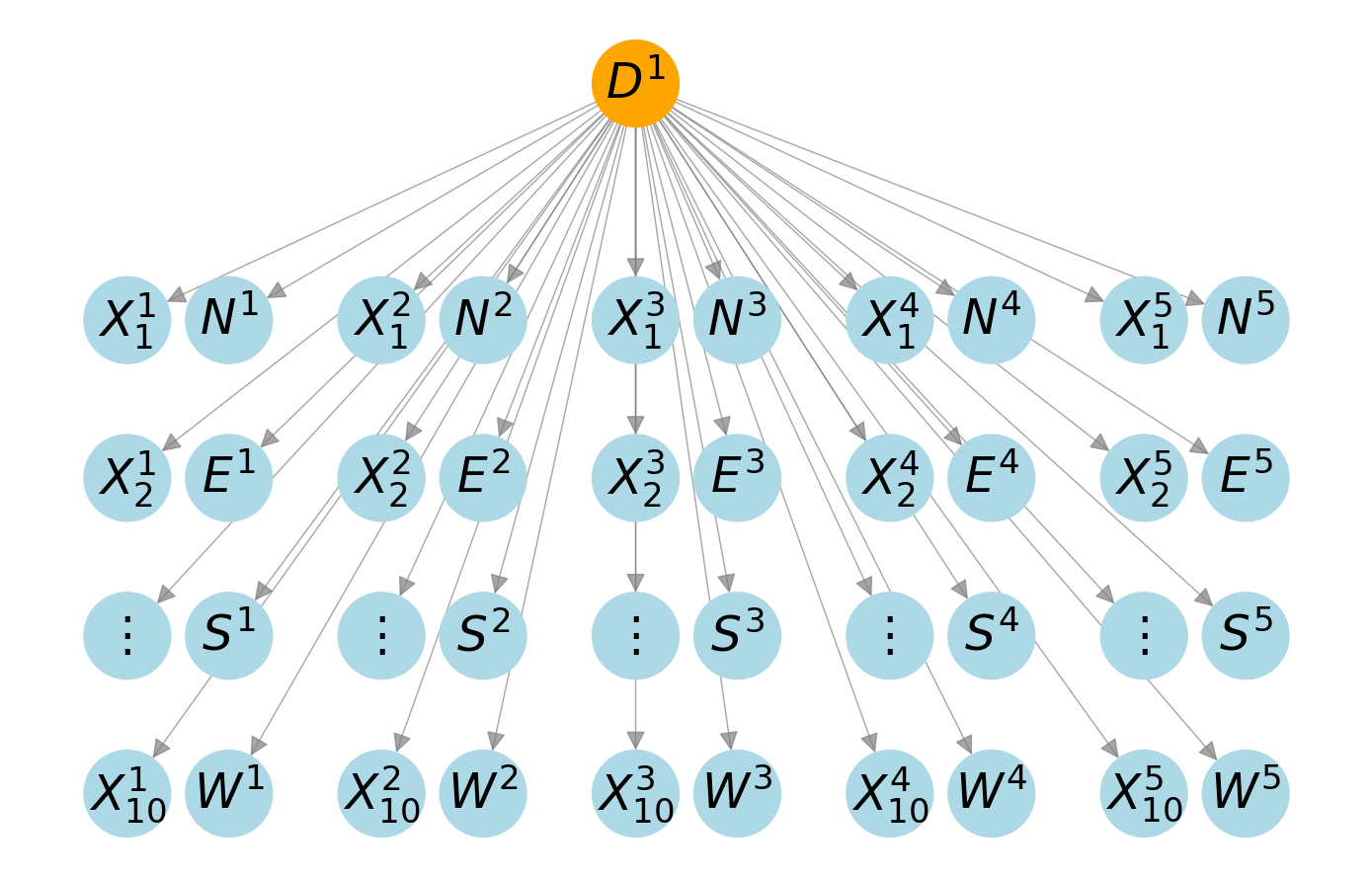}
\includegraphics[scale=0.23]{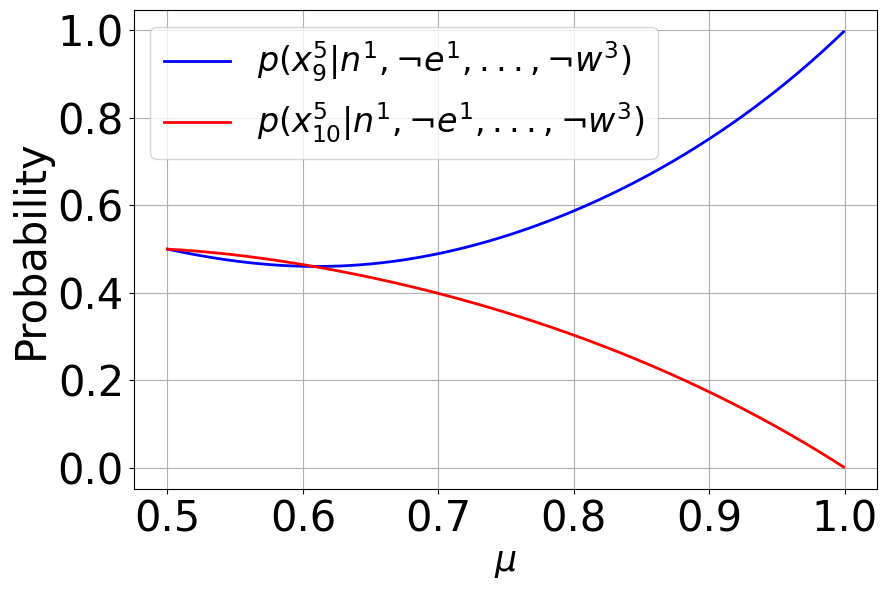}
\caption{$p(x_{i}^{5}|n^{1},\lnot e^{1},\lnot s^{1},\lnot w^{1},n^{2},\lnot e^{2},\lnot s^{2},\lnot w^{2},\lnot n^{3},\lnot e^{3},s^{3},\lnot w^{3})$, for $i\in\{9,10\}$ as a function of $\mu$. Undefined values at $\mu=1$ can be resolved by the limit as $\mu\to 1$.}
\label{fig:hidden_states}
\end{center}
\end{figure}
%%%
\end{example}
%%%%%%%%%%%%%%%%%%%%%%%%%%%%%%%%%%%%%%%%%%%%%%%%%%%%%%%%%%%%%%%%%%%%%%%%%%%%%%%%%%%%%%%%%%%%%%%%%%%%%%%%%%%%%%%%%%%%%%%%%%%%%%%%%%%%%%%%%%%%%%%%%%%%%%%%%%%%%%%%%%
\section{Conclusions}
We formalised a novel temporal probabilistic model of abstraction that enables to reason propositional formulas from data over time. The most practical equations resulting from this research include Equation (\ref{eq:MJD_condition}):
\begin{eqnarray*}
p(\alpha|\Delta)&=&\frac{\sum_{k=1}^{K}\prod_{x_{i}^{t}\in\Delta\cup\{\alpha\}}p(x_{i}^{t}|m(n^{t-1}(d_{k})))}{\sum_{k=1}^{K}\prod_{x_{i}^{t}\in\Delta}p(x_{i}^{t}|m(n^{t-1}(d_{k})))}.
\end{eqnarray*}
This paper focused on the theoretical validations of the probabilistic model. In particular, the model was discussed in terms of data scarcity, a large hypothesis space, and poor data transparency, which are fundamental challenges of standard probabilistic models. We discussed the following points.
\begin{itemize}
\item The above equation derived from our probabilistic model is equivalent to a highest-order, i.e., full-memory, Markov chain trained using maximum likelihood estimation, yet it has essentially linear time complexity with respect to the number of data.
\item Limits provide a natural and mathematically rigorous approach to cope with data scarcity, including the zero-frequency problem.
\item A probability distribution over data generated by our probabilistic model helps data transparency by revealing influential data used in predictions.
\item Our probabilistic model encodes problem-independent inference, in contrast to standard probabilistic models, which encode problem-dependent domain knowledge, such as dependency between hidden and observable variables.
\end{itemize}
Future work includes empirical evaluation of the theoretical advantages of the probabilistic model proposed in this paper.
%%%%%%%%%%%%%%%%%%%%%%%%%%%%%%%%%%%%%%%%%%%%%%%%%%%%%%%%%%%%%%%%%%%%%%%%%%%%%%%%%%%%%%%%%%%%%%%%%%%%%%%%%%%%%%%%%%%%
\bibliographystyle{splncs04}
\bibliography{btx_kido}
%%%%%%%%%%%%%%%%%%%%%%%%%%%%%%%%%%%%%%%%%%%%%%%%%%%%%%%%%%%%%%%%%%%%%%%%%%%%%%%%%%%%%%%%%%%%%%%%%%%%%%%%%%%%%%%%%%%%
\end{document}